\documentclass[runningheads]{llncs}
\usepackage{graphicx}
\usepackage{comment}
\usepackage{amsmath,amssymb} 
\usepackage{color}
\usepackage{multirow}

\newcommand{\etal}{{\it{et al}}.}

\begin{document}
\pagestyle{headings}
\mainmatter
\def\ECCVSubNumber{2186}  

\title{History Repeats Itself: Human Motion Prediction via Motion Attention} 

\titlerunning{History Repeats Itself: Human Motion Prediction via Motion Attention}
%
\author{Wei Mao\inst{1} \and
Miaomiao Liu\inst{1} \and
Mathieu Salzmann\inst{2}}
\authorrunning{W. Mao, M. Liu, M. Salzmann}
%
\institute{Australian National University, Canberra, Australia
\and
EPFL--CVLab \& ClearSpace, Switzerland\\
\email{\{wei.mao,miaomiao.liu\}@anu.edu.au, mathieu.salzmann@epfl.ch}
}
\maketitle

\begin{abstract}
Human motion prediction aims to forecast future human poses given a past motion. Whether based on recurrent or feed-forward neural networks, existing methods fail to model the observation that human motion tends to repeat itself, even for complex sports actions and cooking activities. Here, we introduce an attention-based feed-forward network that explicitly leverages this observation. In particular, instead of modeling frame-wise attention via pose similarity, we propose to extract \emph{motion attention} to capture the similarity between the current motion context and the historical motion sub-sequences. Aggregating the relevant past motions and processing the result with a graph convolutional network allows us to effectively exploit motion patterns from the long-term history to predict the future poses. Our experiments on Human3.6M, AMASS and 3DPW evidence the benefits of our approach for both periodical and non-periodical actions. Thanks to our attention model, it yields state-of-the-art results on all three datasets. Our code is available at \url{https://github.com/wei-mao-2019/HisRepItself}.

\end{abstract}
\section{Introduction}\label{sec:intro}
Human motion prediction consists of forecasting the future poses of a person given a history of their previous motion. Predicting human motion can be highly beneficial for tasks such as human tracking~\cite{gong2011multi}, human-robot interaction~\cite{koppula2013anticipating}, and human motion generation for computer graphics~\cite{2012-ccclde,kovar2008motion,sidenbladh2002implicit}.
To tackle the problem effectively, recent approaches use deep neural networks~\cite{Martinez_2017_CVPR,fragkiadaki2015recurrent,JainZSS16} to model the temporal historical data. 

\begin{figure}[!ht]
    \centering
      \includegraphics[width=\textwidth]{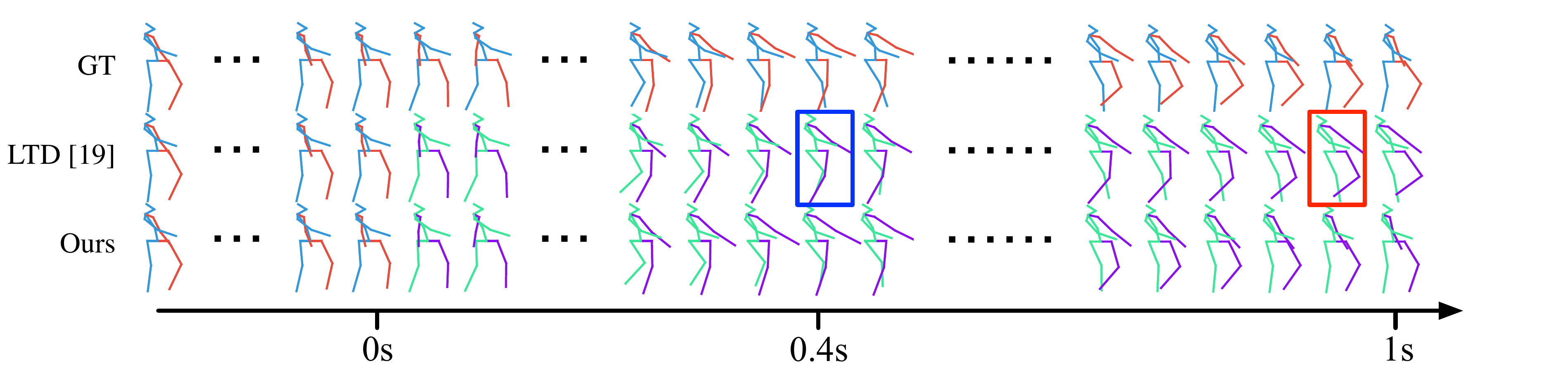}
      \caption{{\bf Human motion prediction} aims to forecast future human poses ($>0s$) given past ones. From top to bottom, we show the ground-truth pose sequence, the predictions of LTD~\cite{mao2019learning} and those of our approach. Frames where LTD~\cite{mao2019learning} makes larger error on arms and legs are highlighted in blue and red box respectively. Note that our results better match the ground truth than those of LTD~\cite{mao2019learning}.}
      \label{fig:intro}
\end{figure}
Traditional methods, such as hidden Markov models~\cite{brand2000style} and Gaussian Process Dynamical Models~\cite{wang2008gaussian}, have proven effective for simple motions, such as walking and golf swings. However, they are typically outperformed by deep learning ones on more complex motions.
The most common trend in modeling the sequential data that constitutes human motion consists of using Recurrent Neural Networks (RNNs)\cite{Martinez_2017_CVPR,fragkiadaki2015recurrent,JainZSS16}. However, as discussed in~\cite{LiZLL18}, in the mid- to long-term horizon, RNNs tend to generate static poses because they struggle to keep track of long-term history. To tackle this problem, existing works~\cite{LiZLL18,gui2018adversarial} either rely on Generative Adversarial Networks (GANs), which are  notoriously hard to train~\cite{ArjovskyB17}, or introduce an additional~\emph{long-term encoder} to represent information from the further past~\cite{LiZLL18}. Unfortunately, such an encoder treats the entire motion history equally, thus not allowing the model to put more emphasis on some parts of the past motion that better reflect the context of the current motion.

In this paper, by contrast, we introduce an attention-based motion prediction approach that effectively exploits historical information by dynamically adapting its focus on the previous motions to the current context. Our method is motivated by the observation that humans tend to repeat their motion, not only in short periodical activities, such as walking, but also in more complex actions occurring across longer time periods, such as sports and cooking activities~\cite{runia2018real}.Therefore, we aim to find the relevant historical information to predict future motion.

To the best of our knowledge, only~\cite{Tang_2018} has attempted to leverage attention for motion prediction. This, however, was achieved in a frame-wise manner, by comparing the human pose from the last observable frame with each one in the historical sequence. As such, this approach fails to reflect the motion direction and is affected by the fact that similar poses may appear in completely different motions. For instance, in most Human3.6M activities, the actor will at some point be standing with their arm resting along their body. To overcome this, we therefore propose to model \emph{motion attention}, and thus compare the last visible sub-sequence with a history of motion sub-sequences.

To this end, inspired by~\cite{mao2019learning}, we represent each sub-sequence in trajectory space using the Discrete Cosine Transform (DCT). We then exploit our motion attention as weights to aggregate the entire DCT-encoded motion history into a future motion estimate. This estimate is combined with the latest observed motion, and the result acts as input to a graph convolutional network (GCN), which lets us better encode spatial dependencies between different joints.As evidenced by our experiments on Human3.6M~\cite{h36m_pami}, AMASS~\cite{AMASS:2019}, and 3DPW~\cite{vonMarcard2018}, and illustrated in Fig.~\ref{fig:intro}, our motion attention-based approach consistently outperforms the state of the art on short-term and long-term motion prediction by training a single unified model for both settings. This contrasts with the previous-best model LTD~\cite{mao2019learning}, which requires training different models for different settings to achieve its best performance.
Furthermore, we demonstrate that it can effectively leverage the repetitiveness of motion in longer sequences.

Our contributions can be summarized as follows. (i) We introduce an attention-based model that exploits motions instead of static frames to better leverage historical information for motion prediction; (ii) Our motion attention allows us to train a unified model for both short-term and long-term prediction; (iii) Our approach can effectively make use of motion repetitiveness in long-term history; (iv) It yields state-of-the-art results and generalizes better than existing methods across datasets and actions.
\section{Related Work}
\noindent{{\bf RNN-based human motion prediction.}} ~RNNs have proven highly successful in sequence-to-sequence prediction tasks~\cite{sutskever2011generating,kiros2015skip}. As such, they have been widely employed for human motion prediction~\cite{fragkiadaki2015recurrent,JainZSS16,Martinez_2017_CVPR,gopalakrishnan2019neural}. 
For instance, Fragkiadaki~\etal~\cite{fragkiadaki2015recurrent} proposed an Encoder-Recurrent-Decoder~(ERD) model that incorporates a non-linear multi-layer feedforward network to encode and decode motion before and after recurrent layers. To avoid error accumulation, curriculum learning was adopted during training. In~\cite{JainZSS16}, Jain~\etal~introduced a Structural-RNN model relying on a manually-designed spatio-temporal graph to encode motion history. 
The fixed structure of this graph, however, restricts the flexibility of this approach at modeling long-range spatial relationships between different limbs. To improve motion estimation,~Martinez \etal~\cite{Martinez_2017_CVPR} proposed a residual-based model that predicts velocities instead of poses. Furthermore, it was shown in this work that a simple zero-velocity baseline, i.e., constantly predicting the last observed pose, led to better performance than~\cite{fragkiadaki2015recurrent,JainZSS16}. While this led to better performance than the previous pose-based methods, the predictions produced by the RNN still suffer from discontinuities between the observed poses and predicted ones.
To overcome this, Gui \etal~proposed to adopt adversarial training to generate smooth sequences~\cite{gui2018adversarial}. In~\cite{hernandez2019human}, Ruiz~\etal~treat human motion prediction as a tensor inpainting problem and exploit a generative adversarial network for long-term prediction. While this approach further improves performance, the use of an adversarial classifier notoriously complicates training~\cite{ArjovskyB17}, making it challenging to deploy on new datasets. 
\begin{figure}[!t]
    \centering
      \includegraphics[width=\textwidth]{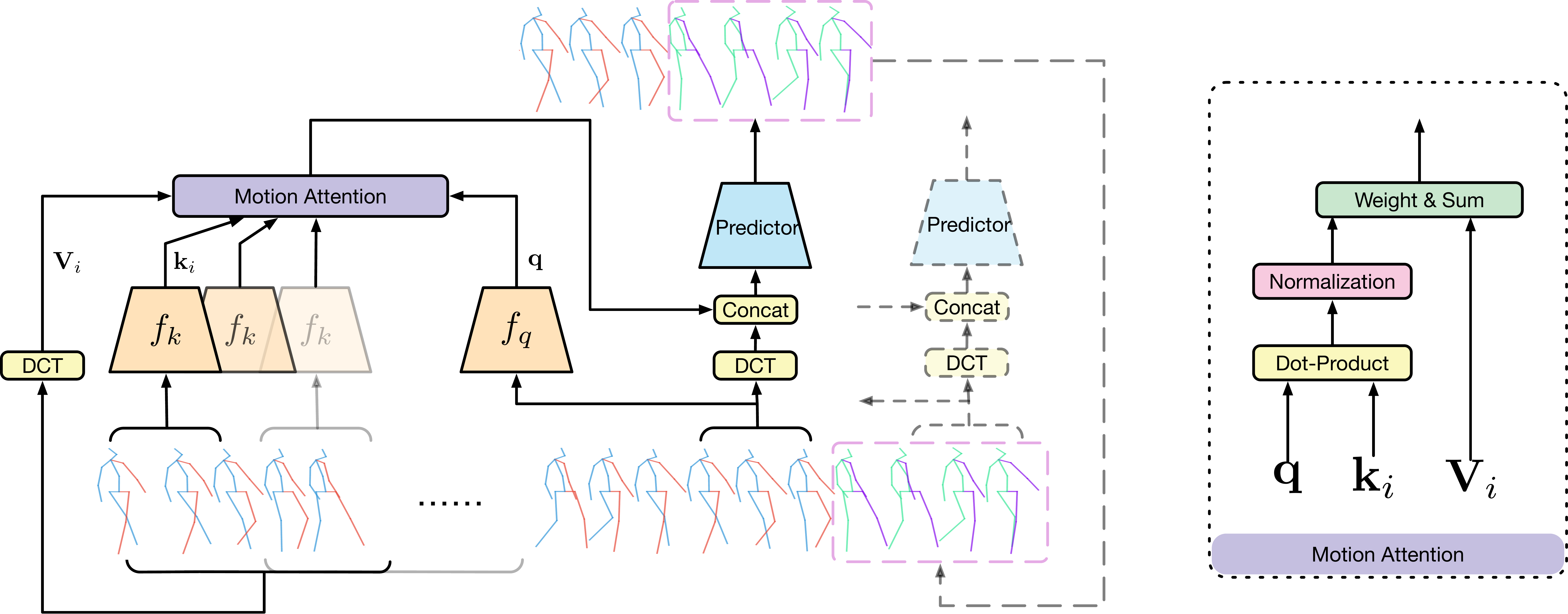}
      \caption{{\bf Overview of our approach.} The past poses are shown as blue and red skeletons and the predicted ones in green and purple. 
      The last observed $M$ poses are initially used as query. 
      For every $M$ consecutive poses in the history (key), we compute an attention score to weigh the DCT coefficients (values) of the corresponding sub-sequence. The weighted sum of such values is then concatenated with the DCT coefficients of the last observed sub-sequence to predict the future. At test time, to predict poses in the further future, we use the output of the predictor as input and predict future motion recursively (as illustrated by the dashed line).}
      \label{fig:net-structure}
\end{figure}

\noindent{{\bf Feed-forward methods and long motion history encoding.}} 
In view of the drawbacks of RNNs, several works considered feed-forward networks as an alternative solution~\cite{Butepage_2017_CVPR,LiZLL18,mao2019learning}. In particular, in~\cite{Butepage_2017_CVPR}, Butepage \etal~introduced a fully-connected network to process the recent pose history, investigating different strategies to encode temporal historical information via convolutions and exploiting the kinematic tree to encode spatial information. However, similar to~\cite{JainZSS16}, and as discussed in~\cite{LiZLL18}, the use of a fixed tree structure does not reflect the motion synchronization across different, potentially distant, human body parts. To capture such dependencies, Li \etal~\cite{LiZLL18}~built a convolutional sequence-to-sequence model processing a two-dimensional pose matrix whose columns represent the pose at every time step. This model was then used to extract a prior from long-term motion history, which, in conjunction with the more recent motion history, was used as input to an autoregressive network for future pose prediction. While more effective than the RNN-based frameworks, the manually-selected size of the convolutional window highly influences the temporal encoding.

Our work is inspired by that of Mao \etal~~\cite{mao2019learning}, who showed that encoding the short-term history in frequency space using the DCT, followed by a GCN to encode spatial and temporal connections led to state-of-the-art performance for human motion prediction up to 1s. As acknowledged by Mao \etal~\cite{mao2019learning}, however, encoding long-term history in DCT yields an overly-general motion representation, leading to worse performance than using short-term history.~In this paper, we overcome this drawback by introducing a~\emph{motion attention} based approach to human motion prediction. This allows us to capture the motion recurrence in the long-term history. Furthermore, in contrast to~\cite{LiZLL18}, whose encoding of past motions depends on the manually-defined size of the temporal convolution filters, our model dynamically adapts its history-based representation to the context of the current prediction.

\noindent{{\bf Attention models for human motion prediction.}}~While attention-based neural networks are commonly employed for machine translation~\cite{vaswani2017attention,bahdanau2014neural}, their use for human motion prediction remains largely unexplored. 
The work of Tang \etal~\cite{Tang_2018} constitutes the only exception, incorporating an attention module to summarize the recent pose history, followed by an RNN-based prediction network. This work, however, uses frame-wise pose-based attention, which may lead to ambiguous motion, because static poses do not provide information about the motion direction and similar poses occur in significantly different motions. To overcome this, we propose to leverage~\emph{motion attention}. As evidenced by our experiments, this, combined with a feed-forward prediction network, allows us to outperform the state-of-the-art motion prediction frameworks.
\section{Our Approach}
Let us now introduce our approach to human motion prediction. Let ${\bf X}_{1:N} = [{\bf x}_1,{\bf x}_2,{\bf x}_3,\cdots,{\bf x}_N]$ encode the motion history, consisting of $N$ consecutive human poses, where ${\bf x}_i\in \mathbb{R}^K$, with $K$ the number of parameters describing each pose, in our case 3D coordinates or angles of human joints.  Our goal is to predict the poses ${\bf X}_{N+1:N+T}$ for the future $T$ time steps. To this end, we introduce a motion attention model that allows us to form a future motion estimate by aggregating the long-term temporal information from the history.
We then combine this estimate with the latest observed motion and input this combination to a GCN-based feed-forward network that lets us learn the spatial and temporal dependencies in the data. 
Below, we discuss these two steps in detail.
\subsection{Motion Attention Model}\label{sec:att-model}
As humans tend to repeat their motion across long time periods, our goal is to discover sub-sequences in the motion history that are similar to the current sub-sequence. In this paper, we propose to achieve this via an attention model.

Following the machine translation formalism of~\cite{vaswani2017attention}, we describe our attention model as a mapping from a~\emph{query} and a set of~\emph{key-value} pairs to an output. The output is a weighted sum of~\emph{values}, where the weight, or \emph{attention}, assigned to each value is a function of its corresponding~\emph{key} and of the~\emph{query}.~Mapping to our motion attention model, the~\emph{query} corresponds to a learned representation of the last observed sub-sequence, and the~\emph{key-value} pairs are treated as a dictionary within which~\emph{keys} are learned representations for historical sub-sequences and~\emph{values} are the corresponding learned future motion representations. Our motion attention model output is defined as the aggregation of these future motion representations based on~\emph{partial motion similarity} between the latest motion sub-sequence and historical sub-sequences.

In our context, we aim to compute attention from short sequences. To this end, we first divide the motion history ${\bf X}_{1:N} = [{\bf x}_1,{\bf x}_2,{\bf x}_3,\cdots,{\bf x}_N]$ into $N-M-T+1$ sub-sequences $\{{\bf X}_{i:i+M+T-1}\}_{i=1}^{N-M-T+1}$, each of which consists of $M+T$ consecutive human poses. By using sub-sequences of length $M+T$, we assume that the predictor, which we will introduce later, exploits the past $M$ frames to predict the future $T$ frames. We then take the first $M$ poses of each sub-sequence ${\bf X}_{i:i+M-1}$ to be a key, and the whole sub-sequence ${\bf X}_{i:i+M+T-1}$ is then the corresponding value. Furthermore, we define the query as the latest sub-sequence ${\bf X}_{N-M+1:N}$ with length $M$. 

To leverage the state-of-the-art representation introduced in~\cite{mao2019learning} and make the output of our attention model consistent with that of the final predictor, we map the resulting values to trajectory space using the DCT on the temporal dimension. That is, we take our final values to be the DCT coefficients $\{{\bf V}_i\}_{i=1}^{N-M-T+1}$, where ${\bf V}_i\in \mathbb{R}^{K \times (M+T)}$. Each row of ${\bf V}_i$ contains the DCT coefficients of one joint coordinate sequence. In practice, we can truncate some high frequencies to avoid predicting jittery motion.

As depicted by Fig.~\ref{fig:net-structure}, the query and keys are used to compute attention scores, which then act as weights to combine the corresponding values. To this end, we first map the query and keys to vectors of the same dimension $d$ by two functions $f_{q}:\mathbb{R}^{K\times M}\rightarrow \mathbb{R}^{d}$ and $f_{k}:\mathbb{R}^{K\times M}\rightarrow \mathbb{R}^{d}$ modeled with neural networks. This can be expressed as
\begin{equation}
    {\bf q} = f_{q}({\bf X}_{N-M+1:N})\;, {\bf k}_{i} = f_{k}({\bf X}_{i:i+M-1})\;
\end{equation}
where ${\bf q},{\bf k}_i\in \mathbb{R}^{d}$, and $i \in \{1,2,\cdots,N-M-T+1\}$.
For each key, we then compute an attention score as
\begin{align}
a_i = \frac{{\bf q}{\bf k}_{i}^T}{\sum_{i=1}^{N-M-T+1}{{\bf q}{\bf k}_{i}^T}}\;.
\end{align}
Note that, instead of the softmax function which is commonly used in attention mechanisms, we simply normalize the attention scores by their sum, which we found to avoid the gradient vanishing problem that may occur when using a softmax. While this division only enforces the sum of the attention scores to be $1$, we further restrict the outputs of $f_q$ and $f_k$ to be non-negative with ReLU~\cite{nair2010rectified} to avoid obtaining negative attention scores.

We then compute the output of the attention model as the weighed sum of values. That is, 
\begin{equation}
    {\bf U} = \sum_{i=1}^{N-M-T+1}{a_i{\bf V}_i}\;,
\end{equation}
where ${\bf U}\in \mathbb{R}^{K \times (M+T)}$. This initial estimate is then combined with the latest sub-sequence and processed by the prediction model described below to generate future poses $\hat{{\bf X}}_{N+1:N+T}$. At test time, to generate longer future motion, we augment the motion history with the last predictions and update the query with the latest sub-sequence in the augmented motion history, and the key-value pairs accordingly. These updated entities are then used for the next prediction step.

\subsection{Prediction Model}\label{sec:pred}
To predict the future motion, we use the state-of-the-art motion prediction model of~\cite{mao2019learning}. Specifically, as mentioned above, we use a DCT-based representation to encode the temporal information for each joint coordinate or angle and GCNs with learnable adjacency matrices to capture the spatial dependencies among these coordinates or angles. 

\begin{table}[ht]
\caption{Short-term prediction of 3D joint positions on H3.6M. The error is measured in millimeter. The two numbers after the method name ``LTD" indicate the number of observed frames and that of future frames to predict, respectively, during training. Our approach achieves state of the art performance across all 15 actions at almost all time horizons, especially for actions with a clear repeated history, such as ``Walking".}
\centering
\resizebox{0.7\textwidth}{!}{%
\begin{tabular}{ccccc|cccc|cccc|cccc}
  & \multicolumn{4}{c}{Walking} & \multicolumn{4}{c}{Eating} & \multicolumn{4}{c}{Smoking} & \multicolumn{4}{c}{Discussion} \\
     milliseconds       & 80   & 160  & 320  & 400  & 80   & 160  & 320  & 400  & 80   & 160  & 320  & 400  & 80   & 160  & 320  & 400   \\\hline
Res. Sup.~\cite{Martinez_2017_CVPR} &  23.2 & 40.9 & 61.0 & 66.1 & 16.8 & 31.5 & 53.5 & 61.7 & 18.9 & 34.7 & 57.5 & 65.4 & 25.7 & 47.8 & 80.0 & 91.3 \\
convSeq2Seq~\cite{LiZLL18} & 17.7 & 33.5 & 56.3 & 63.6 & 11.0 & 22.4 & 40.7 & 48.4 & 11.6 & 22.8 & 41.3 & 48.9 & 17.1 & 34.5 & 64.8 & 77.6  \\
LTD-50-25\cite{mao2019learning} & 12.3 & 23.2 & 39.4 & 44.4 & 7.8 & 16.3 & 31.3 & 38.6 & 8.2 & 16.8 & 32.8 & 39.5 & 11.9 & 25.9 & 55.1 & 68.1 \\
LTD-10-25\cite{mao2019learning} & 12.6 & 23.6 & 39.4 & 44.5 & 7.7 & 15.8 & 30.5 & 37.6 & 8.4 & 16.8 & 32.5 & 39.5 & 12.2 & 25.8 & 53.9 & 66.7 \\
LTD-10-10\cite{mao2019learning} & 11.1 & 21.4 & 37.3 & 42.9 & 7.0 & 14.8 & 29.8 & 37.3 & 7.5 & 15.5 & 30.7 & 37.5 & 10.8 & 24.0 & 52.7 & 65.8 \\\hline
Ours & \textbf{10.0} & \textbf{19.5} & \textbf{34.2} & \textbf{39.8} & \textbf{6.4} & \textbf{14.0} & \textbf{28.7} & \textbf{36.2} & \textbf{7.0} & \textbf{14.9} & \textbf{29.9} & \textbf{36.4} & \textbf{10.2} & \textbf{23.4} & \textbf{52.1} & \textbf{65.4}\\\hline
\end{tabular}
}
\resizebox{1.01\textwidth}{!}{%
\begin{tabular}{ccccc|cccc|cccc|cccc|cccc|cccc}
  & \multicolumn{4}{c}{Directions} & \multicolumn{4}{c}{Greeting} & \multicolumn{4}{c}{Phoning} & \multicolumn{4}{c}{Posing}&\multicolumn{4}{c}{Purchases}&\multicolumn{4}{c}{Sitting} \\
  milliseconds& 80 & 160 & 320 & 400 & 80 & 160 & 320 & 400 & 80 & 160 & 320 & 400 & 80 & 160 & 320 & 400& 80 & 160 & 320 & 400& 80 & 160 & 320 & 400 \\ \hline
  Res. Sup.~\cite{Martinez_2017_CVPR}       & 21.6 & 41.3 & 72.1 & 84.1 & 31.2 & 58.4 & 96.3 & 108.8 & 21.1 & 38.9 & 66.0 & 76.4 & 29.3 & 56.1 & 98.3 & 114.3 & 28.7 & 52.4 & 86.9 & 100.7 & 23.8 & 44.7 & 78.0 & 91.2 \\
convSeq2Seq~\cite{LiZLL18}      & 13.5 & 29.0 & 57.6 & 69.7 & 22.0 & 45.0 & 82.0 & 96.0 & 13.5 & 26.6 & 49.9 & 59.9 & 16.9 & 36.7 & 75.7 & 92.9 & 20.3 & 41.8 & 76.5 & 89.9 & 13.5 & 27.0 & 52.0 & 63.1  \\
LTD-50-25\cite{mao2019learning} & 8.8 & 20.3 & 46.5 & 58.0 & 16.2 & 34.2 & 68.7 & 82.6 & 9.8 & 19.9 & 40.8 & 50.8 & 12.2 & 27.5 & 63.1 & 79.9 & 15.2 & 32.9 & 64.9 & 78.1 & 10.4 & 21.9 & 46.6 & 58.3\\
LTD-10-25\cite{mao2019learning} & 9.2 & 20.6 & 46.9 & 58.8 & 16.7 & 33.9 & 67.5 & 81.6 & 10.2 & 20.2 & 40.9 & 50.9 & 12.5 & 27.5 & 62.5 & 79.6 & 15.5 & 32.3 & 63.6 & 77.3 & 10.4 & 21.4 & 45.4 & 57.3\\
LTD-10-10\cite{mao2019learning} & 8.0 & 18.8 & \textbf{43.7} & \textbf{54.9} & 14.8 & 31.4 & 65.3 & 79.7 & 9.3 & 19.1 & 39.8 & 49.7 & 10.9 & 25.1 & 59.1 & 75.9 & 13.9 & 30.3 & 62.2 & 75.9 & 9.8 & 20.5 & \textbf{44.2} & \textbf{55.9}\\\hline
Ours & \textbf{7.4} & \textbf{18.4} & 44.5 & 56.5 & \textbf{13.7} & \textbf{30.1} & \textbf{63.8} & \textbf{78.1} & \textbf{8.6} & \textbf{18.3} & \textbf{39.0} & \textbf{49.2} & \textbf{10.2} & \textbf{24.2} & \textbf{58.5} & \textbf{75.8} & \textbf{13.0} & \textbf{29.2} & \textbf{60.4} & \textbf{73.9} & \textbf{9.3} & \textbf{20.1} & 44.3 & 56.0 \\ \hline
  & \multicolumn{4}{c}{Sitting Down} & \multicolumn{4}{c}{Taking Photo} & \multicolumn{4}{c}{Waiting} & \multicolumn{4}{c}{Walking Dog}&\multicolumn{4}{c}{Walking Together}&\multicolumn{4}{c}{Average} \\
  milliseconds& 80 & 160 & 320 & 400 & 80 & 160 & 320 & 400 & 80 & 160 & 320 & 400 & 80 & 160 & 320 & 400& 80 & 160 & 320 & 400& 80 & 160 & 320 & 400 \\ \hline
  Res. Sup.~\cite{Martinez_2017_CVPR} & 31.7 & 58.3 & 96.7 & 112.0 & 21.9 & 41.4 & 74.0 & 87.6 & 23.8 & 44.2 & 75.8 & 87.7 & 36.4 & 64.8 & 99.1 & 110.6 & 20.4 & 37.1 & 59.4 & 67.3 & 25.0 & 46.2 & 77.0 & 88.3 \\
convSeq2Seq~\cite{LiZLL18} & 20.7 & 40.6 & 70.4 & 82.7 & 12.7 & 26.0 & 52.1 & 63.6 & 14.6 & 29.7 & 58.1 & 69.7 & 27.7 & 53.6 & 90.7 & 103.3 & 15.3 & 30.4 & 53.1 & 61.2 & 16.6 & 33.3 & 61.4 & 72.7  \\
LTD-50-25\cite{mao2019learning} & 17.1 & 34.2 & 63.6 & 76.4 & 9.6 & 20.3 & 43.3 & 54.3 & 10.4 & 22.1 & 47.9 & 59.2 & 22.8 & 44.7 & 77.2 & 88.7 & 10.3 & 21.2 & 39.4 & 46.3 & 12.2 & 25.4 & 50.7 & 61.5\\
LTD-10-25\cite{mao2019learning} & 17.0 & 33.4 & 61.6 & 74.4 & 9.9 & 20.5 & 43.8 & 55.2 & 10.5 & 21.6 & 45.9 & 57.1 & 22.9 & 43.5 & 74.5 & 86.4 & 10.8 & 21.7 & 39.6 & 47.0 & 12.4 & 25.2 & 49.9 & 60.9\\
LTD-10-10\cite{mao2019learning} & 15.6 & 31.4 & \textbf{59.1} & \textbf{71.7} & 8.9 & 18.9 & 41.0 & 51.7 & 9.2 & 19.5 & \textbf{43.3} & \textbf{54.4} & 20.9 & 40.7 & 73.6 & 86.6 & 9.6 & 19.4 & 36.5 & 44.0 & 11.2 & 23.4 & 47.9 & 58.9\\\hline
Ours & \textbf{14.9} & \textbf{30.7} & \textbf{59.1} & 72.0 & \textbf{8.3} & \textbf{18.4} & \textbf{40.7} & \textbf{51.5} & \textbf{8.7} & \textbf{19.2} & 43.4 & 54.9 & \textbf{20.1} & \textbf{40.3} & \textbf{73.3} & \textbf{86.3} & \textbf{8.9} & \textbf{18.4} & \textbf{35.1} & \textbf{41.9} & \textbf{10.4} & \textbf{22.6} & \textbf{47.1} & \textbf{58.3} \\\hline
\end{tabular}
}
\label{tab:h36_short_3d}
\end{table}

\noindent{{\bf Temporal encoding}.} Given a sequence of $k^{th}$ joint coordinates or angles $\{x_{k,l}\}_{l=1}^{L}$ or its DCT coefficients $\{C_{k,l}\}_{l=1}^{L}$, the DCT and Inverse-DCT~(IDCT) are,
\begin{equation}
\resizebox{0.93\textwidth}{!}{
$C_{k,l} = \sqrt{\frac{2}{L}}\sum\limits_{n=1}^{L}x_{k,n}\frac{1}{\sqrt{1+\delta_{l1}}}\cos\left(\frac{\pi}{L}(n-\frac{1}{2})(l-1)\right)\;,x_{k,n} =\sqrt{\frac{2}{L}}\sum\limits_{l=1}^{L}C_{k,l}\frac{1}{\sqrt{1+\delta_{l1}}}\cos\left(\frac{\pi}{L}(n-\frac{1}{2})(l-1)\right)\;$
}
\end{equation}
where $l\in\{1,2,\cdots,L\}$,$n\in\{1,2,\cdots,L\}$ and $\delta_{ij}=\begin{cases}
  1 & \text{if}\ i=j\\
  0 & \text{if}\ i\neq j.
  \end{cases}$.

To predict future poses ${\bf X}_{N+1:N+T}$, we make use of the latest sub-sequence ${\bf X}_{N-M+1:N}$, which is also the query in the attention model. Adopting the same padding strategy as~\cite{mao2019learning},  we then replicate the last observed pose ${\bf X}_N$ $T$ times to generate a sequence of length $M+T$ and the DCT coefficients of this sequence are denoted as ${\bf D}\in \mathbb{R}^{K\times (M+T)}$. We then aim to predict DCT coefficients of the future sequence ${\bf X}_{N-M+1:N+T}$ given ${\bf D}$ and the attention model's output ${\bf U}$.

\noindent{{\bf Spatial encoding}.} To capture spatial dependencies between different joint coordinates or angles, we regard the human body as a fully-connected graph with $K$ nodes. 
The input to a graph convolutional layer $p$ is a matrix ${\bf H}^{(p)}\in \mathbb{R}^{K\times F}$, where each row is the $F$ dimensional feature vector of one node. For example, for the first layer, the network takes as input the $K \times 2(M+T)$ matrix that concatenates ${\bf D}$ and ${\bf U}$. A graph convolutional layer then outputs a $K\times\hat{F}$ matrix of the form
\begin{equation}
    {\bf H}^{(p+1)} = \sigma({\bf A}^{(p)}{\bf H}^{(p)}{\bf W}^{(p)})\;,
\end{equation}
where ${\bf A}^{(p)}\in \mathbb{R}^{K\times K}$ is the trainable adjacency matrix of layer $p$, representing the strength of the connectivity between nodes, ${\bf W}^{(p)}\in \mathbb{R}^{F \times \hat{F}}$ also encodes trainable weights but used to extract features, and $\sigma(\cdot)$ is an activation function, such as $tanh(\cdot)$. We stack several such layers to form our GCN-based predictor.

Given ${\bf D}$ and ${\bf U}$, the predictor learns a residual between the DCT coefficients ${\bf D}$ of the padded sequence and those of the true sequence. By applying IDCT to the predicted DCT coefficients, we obtain the coordinates or angles $\hat{{\bf X}}_{N-M+1:N+T}$, whose last $T$ poses $\hat{{\bf X}}_{N+1:N+T}$ are predictions in the future.
\begin{table}[ht]
\caption{Long-term prediction of 3D joint positions on H3.6M. On average, our approach performs the best. }
\centering
\resizebox{0.7\textwidth}{!}{%
\begin{tabular}{ccccc|cccc|cccc|cccc}
  & \multicolumn{4}{c}{Walking} & \multicolumn{4}{c}{Eating} & \multicolumn{4}{c}{Smoking} & \multicolumn{4}{c}{Discussion} \\
 milliseconds & 560 & 720 & 880 & 1000 & 560 & 720 & 880 & 1000 & 560 & 720 & 880 & 1000 & 560 & 720 & 880 & 1000\\\hline
Res. Sup.~\cite{Martinez_2017_CVPR} & 71.6 & 72.5 & 76.0 & 79.1 & 74.9 & 85.9 & 93.8 & 98.0 & 78.1 & 88.6 & 96.6 & 102.1 & 109.5 & 122.0 & 128.6 & 131.8\\
convSeq2Seq~\cite{LiZLL18} & 72.2 & 77.2 & 80.9 & 82.3 & 61.3 & 72.8 & 81.8 & 87.1 & 60.0 & 69.4 & 77.2 & 81.7 & 98.1 & 112.9 & 123.0 & 129.3\\
LTD-50-25\cite{mao2019learning} & 50.7 & 54.4 & 57.4 & 60.3 & 51.5 & 62.6 & 71.3 & 75.8 & 50.5 & 59.3 & 67.1 & 72.1 & 88.9 & 103.9 & 113.6 & \textbf{118.5}\\
LTD-10-25\cite{mao2019learning} & 51.8 & 56.2 & 58.9 & 60.9 & \textbf{50.0} & \textbf{61.1} & \textbf{69.6} & \textbf{74.1} & 51.3 & 60.8 & 68.7 & 73.6 & 87.6 & 103.2 & \textbf{113.1} & 118.6\\
LTD-10-10\cite{mao2019learning} & 53.1 & 59.9 & 66.2 & 70.7 & 51.1 & 62.5 & 72.9 & 78.6 & 49.4 & 59.2 & 66.9 & 71.8 & 88.1 & 104.5 & 115.5 & 121.6\\\hline
Ours & \textbf{47.4} & \textbf{52.1} & \textbf{55.5} & \textbf{58.1} & \textbf{50.0} & 61.4 & 70.6 & 75.7 & \textbf{47.6} & \textbf{56.6} & \textbf{64.4} & \textbf{69.5} & \textbf{86.6} & \textbf{102.2} & 113.2 & 119.8\\ \hline
\end{tabular}
}

\resizebox{1.01\textwidth}{!}{%
\begin{tabular}{ccccc|cccc|cccc|cccc|cccc|cccc}
  & \multicolumn{4}{c}{Directions} & \multicolumn{4}{c}{Greeting} & \multicolumn{4}{c}{Phoning} & \multicolumn{4}{c}{Posing}&\multicolumn{4}{c}{Purchases}&\multicolumn{4}{c}{Sitting} \\
  milliseconds& 560 & 720 & 880 & 1000 & 560 & 720 & 880 & 1000 & 560 & 720 & 880 & 1000 & 560 & 720 & 880 & 1000 & 560 & 720 & 880 & 1000 & 560 & 720 & 880 & 1000\\\hline
Res. Sup.~\cite{Martinez_2017_CVPR} & 101.1 & 114.5 & 124.5 & 129.1 & 126.1 & 138.8 & 150.3 & 153.9 & 94.0 & 107.7 & 119.1 & 126.4 & 140.3 & 159.8 & 173.2 & 183.2 & 122.1 & 137.2 & 148.0 & 154.0 & 113.7 & 130.5 & 144.4 & 152.6\\
convSeq2Seq~\cite{LiZLL18} & 86.6 & 99.8 & 109.9 & 115.8 & 116.9 & 130.7 & 142.7 & 147.3 & 77.1 & 92.1 & 105.5 & 114.0 & 122.5 & 148.8 & 171.8 & 187.4 & 111.3 & 129.1 & 143.1 & 151.5 & 82.4 & 98.8 & 112.4 & 120.7\\
LTD-50-25\cite{mao2019learning} & 74.2 & 88.1 & 99.4 & \textbf{105.5} & 104.8 & 119.7 & \textbf{132.1} & \textbf{136.8} & 68.8 & 83.6 & 96.8 & 105.1 & 110.2 & 137.8 & 160.8 & 174.8 & 99.2 & 114.9 & 127.1 & 134.9 & 79.2 & 96.2 & 110.3 & 118.7\\
LTD-10-25\cite{mao2019learning} & 76.1 & 91.0 & 102.8 & 108.8 & 104.3 & 120.9 & 134.6 & 140.2 & 68.7 & 84.0 & 97.2 & 105.1 & 109.9 & 136.8 & \textbf{158.3} & \textbf{171.7} & 99.4 & 114.9 & 127.9 & 135.9 & 78.5 & 95.7 & 110.0 & 118.8\\
LTD-10-10\cite{mao2019learning} & \textbf{72.2} & \textbf{86.7} & \textbf{98.5} & 105.8 & 103.7 & 120.6 & 134.7 & 140.9 & 67.8 & 83.0 & \textbf{96.4} & 105.1 & \textbf{107.6} & \textbf{136.1} & 159.5 & 175.0 & 98.3 & 115.1 & 130.1 & 139.3 & \textbf{76.4} & \textbf{93.1} & \textbf{106.9} & \textbf{115.7}\\\hline
Ours & 73.9 & 88.2 & 100.1 & 106.5 & \textbf{101.9} & \textbf{118.4} & 132.7 & 138.8 & \textbf{67.4} & \textbf{82.9} & 96.5 & \textbf{105.0} & \textbf{107.6} & 136.8 & 161.4 & 178.2 & \textbf{95.6} & \textbf{110.9} & \textbf{125.0} & \textbf{134.2} & \textbf{76.4} & \textbf{93.1} & 107.0 & 115.9 \\ \hline
  & \multicolumn{4}{c}{Sitting Down} & \multicolumn{4}{c}{Taking Photo} & \multicolumn{4}{c}{Waiting} & \multicolumn{4}{c}{Walking Dog}&\multicolumn{4}{c}{Walking Together}&\multicolumn{4}{c}{Average} \\
  milliseconds & 560 & 720 & 880 & 1000 & 560 & 720 & 880 & 1000 & 560 & 720 & 880 & 1000 & 560 & 720 & 880 & 1000 & 560 & 720 & 880 & 1000 & 560 & 720 & 880 & 1000 \\\hline
Res. Sup.~\cite{Martinez_2017_CVPR} & 138.8 & 159.0 & 176.1 & 187.4 & 110.6 & 128.9 & 143.7 & 153.9 & 105.4 & 117.3 & 128.1 & 135.4 & 128.7 & 141.1 & 155.3 & 164.5 & 80.2 & 87.3 & 92.8 & 98.2 & 106.3 & 119.4 & 130.0 & 136.6 \\
convSeq2Seq~\cite{LiZLL18} & 106.5 & 125.1 & 139.8 & 150.3 & 84.4 & 102.4 & 117.7 & 128.1 & 87.3 & 100.3 & 110.7 & 117.7 & 122.4 & 133.8 & 151.1 & 162.4 & 72.0 & 77.7 & 82.9 & 87.4 & 90.7 & 104.7 & 116.7 & 124.2 \\
LTD-50-25\cite{mao2019learning} & 100.2 & 118.2 & 133.1 & 143.8 & 75.3 & 93.5 & 108.4 & 118.8 & 77.2 & 90.6 & 101.1 & 108.3 & 107.8 & 120.3 & 136.3 & 146.4 & 56.0 & 60.3 & 63.1 & 65.7 & 79.6 & 93.6 & 105.2 & 112.4 \\
LTD-10-25\cite{mao2019learning} & 99.5 & 118.5 & 133.6 & 144.1 & 76.8 & 95.3 & 110.3 & 120.2 & 75.1 & 88.7 & \textbf{99.5} & \textbf{106.9} & \textbf{105.8} & \textbf{118.7} & \textbf{132.8} & \textbf{142.2} & 58.0 & 63.6 & 67.0 & 69.6 & 79.5 & 94.0 & 105.6 & 112.7 \\
LTD-10-10\cite{mao2019learning} & \textbf{96.2} & \textbf{115.2} & \textbf{130.8} & \textbf{142.2} & 72.5 & 90.9 & 105.9 & \textbf{116.3} & \textbf{73.4} & \textbf{88.2} & 99.8 & 107.5 & 109.7 & 122.8 & 139.0 & 150.1 & 55.7 & 61.3 & 66.4 & 69.8 & 78.3 & 93.3 & 106.0 & 114.0 \\\hline
Ours & 97.0 & 116.1 & 132.1 & 143.6 & \textbf{72.1} & \textbf{90.4} & \textbf{105.5} & 115.9 & 74.5 & 89.0 & 100.3 & 108.2 & 108.2 & 120.6 & 135.9 & 146.9 & \textbf{52.7} & \textbf{57.8} & \textbf{62.0} & \textbf{64.9} & \textbf{77.3} & \textbf{91.8} & \textbf{104.1} & \textbf{112.1} \\\hline
\end{tabular}
}
\label{tab:h36_long_3d}
\end{table}
\subsection{Training}\label{sec:train}
Let us now introduce the loss functions we use to train our model on either 3D coordinates or joint angles.
For 3D joint coordinates prediction, following~\cite{mao2019learning}, we make use of the Mean Per Joint Position Error (MPJPE) proposed in~\cite{h36m_pami}. In particular, for one training sample, this yields the loss
\begin{equation}
    \ell = \frac{1}{J(M+T)}\sum_{t=1}^{M+T}\sum_{j=1}^{J}\|\hat{\textbf{p}}_{t,j}-\textbf{p}_{t,j}\|^2\;,
\end{equation}
where $\hat{\textbf{p}}_{t,j}\in \mathbb{R}^3$ represents the 3D coordinates of the $j^{th}$ joint of the $t^{th}$ human pose in $\hat{{\bf X}}_{N-M+1:N+T}$, and $\textbf{p}_{t,j}\in \mathbb{R}^3$ is the corresponding ground truth.

For the angle-based representation, we use the average $\ell_1$ distance between the predicted joint angles and the ground truth as loss. For one sample, this can be expressed as
\begin{equation}
    \ell = \frac{1}{K(M+T)}\sum_{t=1}^{M+T}\sum_{k=1}^{K}|\hat{x}_{t,k}-x_{t,k}|\;,
\end{equation}
where $\hat{x}_{t,k}$ is the predicted $k^{th}$ angle of the $t^{th}$ pose in $\hat{{\bf X}}_{N-M+1:N+T}$ and $x_{t,k}$ is the corresponding ground truth.
\begin{figure}[ht]
    \centering
    \begin{tabular}{cc}
      \includegraphics[width=0.46\linewidth]{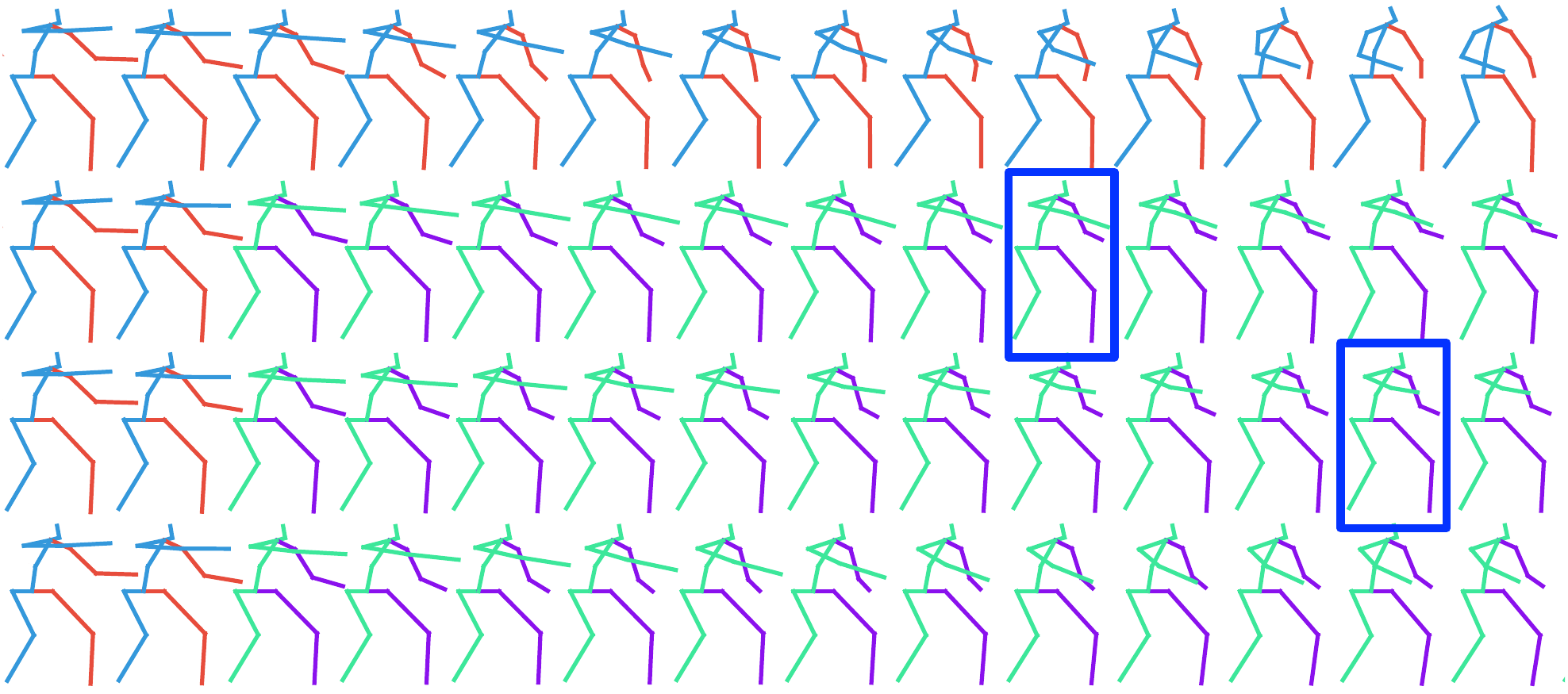} & 
      \includegraphics[width=0.46\linewidth]{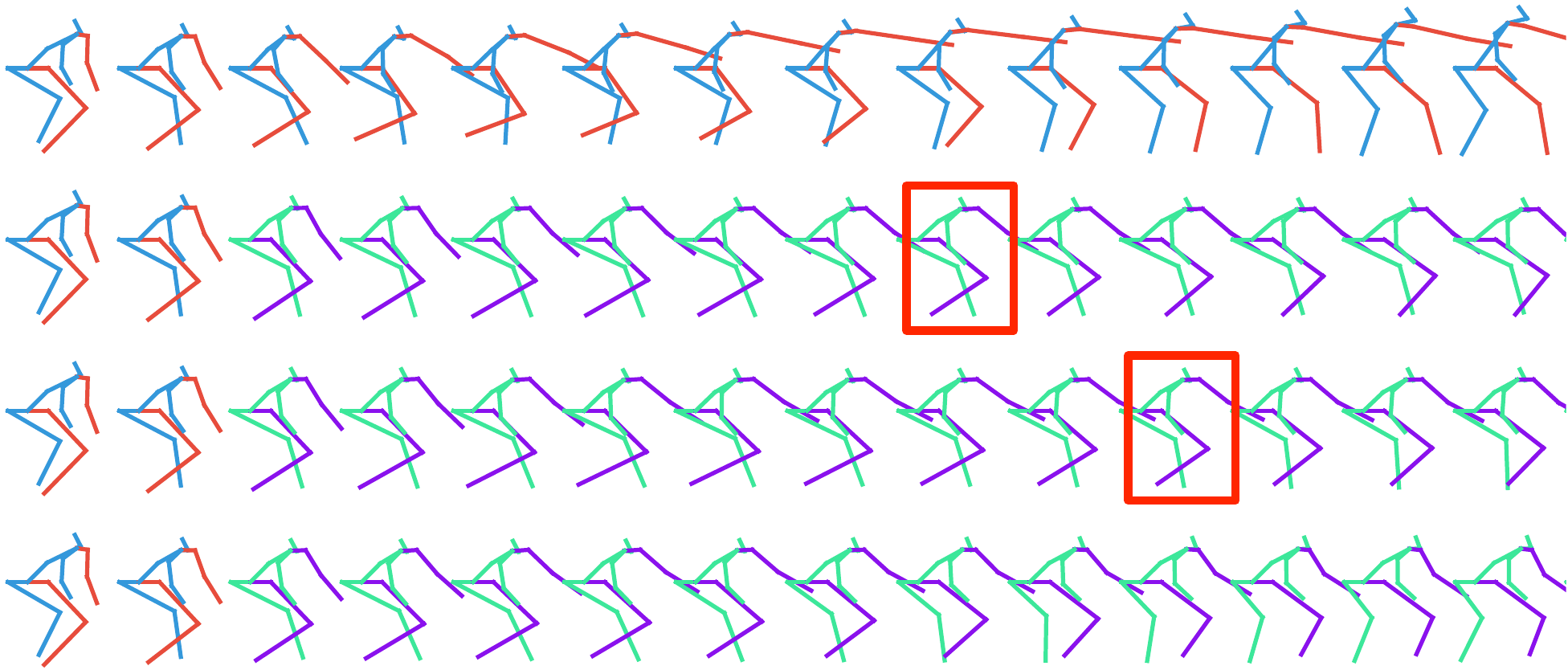} \\
      (a)Discussion & (b)Walking Dog\\
      \multicolumn{2}{c}{\includegraphics[width=0.96\linewidth]{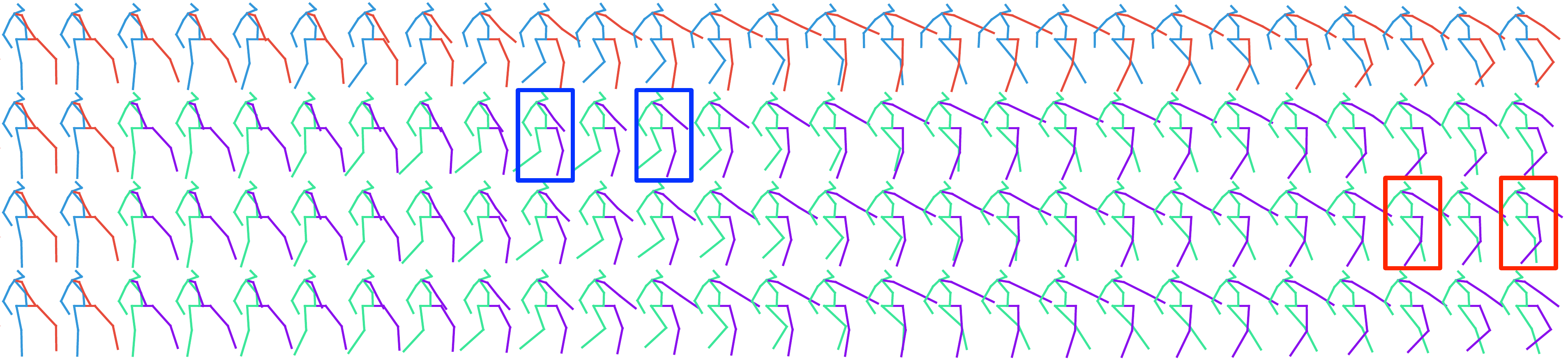}}\\
      \multicolumn{2}{c}{(c)Walking}
    \end{tabular}
    \caption{Qualitative comparison of short-term (``Discussion" and ``Walking Dog") and long-term (``Walking") predictions on H3.6M. From top to bottom, we show the ground truth, and the results of LTD-10-25, LTD-10-10 and our approach on 3D positions. The ground truth is shown as blue-red skeletons, and the predictions as green-purple ones.} 
    \label{fig:short-pred-quali-h36m}
\end{figure}
\subsection{Network Structure}
As shown in Fig.~\ref{fig:net-structure}, our complete framework consists of two modules: a motion attention model and a predictor. For the attention model, we use the same architecture for $f_q$ and $f_k$. Specifically, we use a network consisting of two 1D convolutional layers, each of which is followed by a ReLU~\cite{nair2010rectified} activation function. In our experiments, the kernel size of these two layers is 6 and 5, respectively, to obtain a receptive field of 10 frames. The dimension of the hidden features, the query vector $\textbf{q}$ and the key vectors $\{\textbf{k}_i\}_{i=1}^{N-M-T+1}$ is set to 256.

For the predictor, we use the same GCN with residual structure as in~\cite{mao2019learning}. It is made of 12 residual blocks, each of which contains two graph convolutional layers, with an additional initial layer to map the DCT coefficients to features and a final layer to decode the features to DCT residuals. The learnable weight matrix $\textbf{W}$ of each layer is of size $256\times 256$, and the size of the learnable adjacency matrix $\textbf{A}$ depends on the dimension of one human pose. For example, for 3D coordinates, $\textbf{A}$ is of size $66\times66$. Thanks to the simple structure of our attention model, the overall network remains still compact. Specifically, in our experiments, it has around 3.4 million parameters for both 3D coordinates and angles. The implementation details are included in supplementary material.
\section{Experiments}
Following previous works~\cite{gopalakrishnan2019neural,LiZLL18,mao2019learning,Martinez_2017_CVPR,pavllo2019modeling}, we evaluate our method on Human3.6m (H3.6M)~\cite{h36m_pami} and AMASS~\cite{AMASS:2019}. We further evaluate our method on 3DPW~\cite{vonMarcard2018} using our model trained on AMASS to demonstrate the generalizability of our approach. Below, we discuss these datasets, the evaluation metric and the baseline methods, and present our results using joint angles and 3D coordinates.
\begin{table}[ht]
\caption{Short-term prediction of joint angles on H3.6M. Note that, we report results on 8 sub-sequences per action for fair comparison with MHU~\cite{Tang_2018}.}
\centering
\resizebox{0.7\textwidth}{!}{%
\begin{tabular}{ccccc|cccc|cccc|cccc}
  & \multicolumn{4}{c}{Walking} & \multicolumn{4}{c}{Eating} & \multicolumn{4}{c}{Smoking} & \multicolumn{4}{c}{Discussion} \\
     milliseconds       & 80   & 160  & 320  & 400  & 80   & 160  & 320  & 400  & 80   & 160  & 320  & 400  & 80   & 160  & 320  & 400   \\\hline
Res. sup.~\cite{Martinez_2017_CVPR}& 0.28 & 0.49 & 0.72 & 0.81 & 0.23 & 0.39 & 0.62 & 0.76 & 0.33 & 0.61 & 1.05 & 1.15 & 0.31 & 0.68 & 1.01 & 1.09 \\
convSeq2Seq~\cite{LiZLL18} & 0.33 & 0.54 & 0.68 & 0.73 & 0.22 & 0.36 & 0.58 & 0.71 & 0.26 & 0.49 & 0.96 & 0.92 & 0.32 & 0.67 & 0.94 & 1.01 \\
MHU~\cite{Tang_2018} & 0.32 & 0.53 & 0.69 & 0.77 & - & - & - & - & - & - & - & - & 0.31 & 0.66 & 0.93 & 1.00 \\
LTD-10-25~\cite{mao2019learning} & 0.20 & 0.34 & 0.52 & 0.59 & 0.17 & 0.31 & 0.52 & 0.64 & 0.23 & 0.42 & 0.85 & 0.80 & 0.22 & 0.58 & 0.87 & 0.96 \\
LTD-10-10~\cite{mao2019learning} & \textbf{0.18} & 0.31 & 0.49 & 0.56 & \textbf{0.16} & \textbf{0.29} & 0.50 & 0.62 & \textbf{0.22} & \textbf{0.41} & \textbf{0.86} & \textbf{0.80} & \textbf{0.20} & \textbf{0.51} & \textbf{0.77} & \textbf{0.85} \\\hline
Ours & \textbf{0.18} & \textbf{0.30} & \textbf{0.46} & \textbf{0.51} & \textbf{0.16} & \textbf{0.29} & \textbf{0.49} & \textbf{0.60} & \textbf{0.22} & 0.42 & \textbf{0.86} & \textbf{0.80} & \textbf{0.20} & 0.52 & 0.78 & 0.87\\\hline
\end{tabular}
}
\resizebox{\textwidth}{!}{%
\begin{tabular}{ccccc|cccc|cccc|cccc|cccc|cccc}
  & \multicolumn{4}{c}{Directions} & \multicolumn{4}{c}{Greeting} & \multicolumn{4}{c}{Phoning} & \multicolumn{4}{c}{Posing}&\multicolumn{4}{c}{Purchases}&\multicolumn{4}{c}{Sitting} \\
  milliseconds& 80 & 160 & 320 & 400 & 80 & 160 & 320 & 400 & 80 & 160 & 320 & 400 & 80 & 160 & 320 & 400& 80 & 160 & 320 & 400& 80 & 160 & 320 & 400 \\ \hline
Res. sup.~\cite{Martinez_2017_CVPR}& 0.26 & 0.47 & 0.72 & 0.84 & 0.75 & 1.17 & 1.74 & 1.83 & 0.23 & 0.43 & 0.69 & 0.82 & 0.36 & 0.71 & 1.22 & 1.48 & 0.51 & 0.97 & 1.07 & 1.16 & 0.41 & 1.05 & 1.49 & 1.63 \\
convSeq2Seq~\cite{LiZLL18} & 0.39 & 0.60 & 0.80 & 0.91 & 0.51 & 0.82 & 1.21 & 1.38 & 0.59 & 1.13 & 1.51 & 1.65 & 0.29 & 0.60 & 1.12 & 1.37 & 0.63 & 0.91 & 1.19 & 1.29 & 0.39 & 0.61 & 1.02 & 1.18 \\
MHU~\cite{Tang_2018} & - & - & - & - & 0.54 & 0.87 & 1.27 & 1.45 & - & - & - & - & 0.33 & 0.64 & 1.22 & 1.47 & - & - & - & - & - & - & - & - \\
LTD-10-25~\cite{mao2019learning} & 0.29 & 0.47 & 0.69 & 0.76 & 0.36 & 0.61 & 0.97 & 1.14 & 0.54 & 1.03 & 1.34 & 1.47 & 0.21 & 0.47 & 1.07 & 1.31 & 0.50 & 0.72 & 1.06 & 1.12 & 0.31 & 0.46 & \textbf{0.79} & \textbf{0.95} \\
LTD-10-10~\cite{mao2019learning} & 0.26 & 0.45 & 0.71 & 0.79 & 0.36 & \textbf{0.60} & \textbf{0.95} & \textbf{1.13} & \textbf{0.53} & 1.02 & 1.35 & 1.48 & \textbf{0.19} & \textbf{0.44} & \textbf{1.01} & \textbf{1.24} & 0.43 & \textbf{0.65} & 1.05 & 1.13 & \textbf{0.29} & \textbf{0.45} & 0.80 & 0.97 \\\hline
Ours & \textbf{0.25} & \textbf{0.43} & \textbf{0.60} & \textbf{0.69} & \textbf{0.35} & \textbf{0.60} & \textbf{0.95} & 1.14 & \textbf{0.53} & \textbf{1.01} & \textbf{1.31} & \textbf{1.43} & \textbf{0.19} & 0.46 & 1.09 & 1.35 & \textbf{0.42} & \textbf{0.65} & \textbf{1.00} & \textbf{1.07} & \textbf{0.29} & 0.47 & 0.83 & 1.01 \\\hline
  & \multicolumn{4}{c}{Sitting Down} & \multicolumn{4}{c}{Taking Photo} & \multicolumn{4}{c}{Waiting} & \multicolumn{4}{c}{Walking Dog}&\multicolumn{4}{c}{Walking Together}&\multicolumn{4}{c}{Average} \\
  milliseconds& 80 & 160 & 320 & 400 & 80 & 160 & 320 & 400 & 80 & 160 & 320 & 400 & 80 & 160 & 320 & 400& 80 & 160 & 320 & 400& 80 & 160 & 320 & 400 \\ \hline
Res. sup.~\cite{Martinez_2017_CVPR}& 0.39 & 0.81 & 1.40 & 1.62 & 0.24 & 0.51 & 0.90 & 1.05 & 0.28 & 0.53 & 1.02 & 1.14 & 0.56 & 0.91 & 1.26 & 1.40 & 0.31 & 0.58 & 0.87 & 0.91 & 0.36 & 0.67 & 1.02 & 1.15 \\
convSeq2Seq~\cite{LiZLL18} & 0.41 & 0.78 & 1.16 & 1.31 & 0.23 & 0.49 & 0.88 & 1.06 & 0.30 & 0.62 & 1.09 & 1.30 & 0.59 & 1.00 & 1.32 & 1.44 & 0.27 & 0.52 & 0.71 & 0.74 & 0.38 & 0.68 & 1.01 & 1.13 \\
MHU~\cite{Tang_2018} & - & - & - & - & 0.27 & 0.54 & 0.84 & 0.96 & - & - & - & - & 0.56 & 0.88 & 1.21 & 1.37 & - & - & - & - & 0.39 & 0.68 & 1.01 & 1.13 \\
LTD-10-25~\cite{mao2019learning} & 0.31 & 0.64 & 0.94 & 1.07 & 0.17 & 0.38 & 0.62 & 0.74 & 0.25 & 0.52 & 0.96 & 1.17 & 0.49 & 0.80 & 1.11 & 1.26 & 0.18 & 0.39 & 0.56 & 0.63 & 0.30 & 0.54 & 0.86 & 0.97 \\
LTD-10-10~\cite{mao2019learning} & 0.30 & \textbf{0.61} & \textbf{0.90} & \textbf{1.00} & \textbf{0.14} & \textbf{0.34} & \textbf{0.58} & \textbf{0.70} & 0.23 & 0.50 & \textbf{0.91} & \textbf{1.14} & \textbf{0.46} & 0.79 & 1.12 & 1.29 & 0.15 & 0.34 & 0.52 & 0.57 & \textbf{0.27} & \textbf{0.52} & 0.83 & 0.95 \\\hline
Ours & \textbf{0.30} & 0.63 & 0.92 & 1.04 & 0.16 & 0.36 & \textbf{0.58} & \textbf{0.70} & \textbf{0.22} & \textbf{0.49} & 0.92 & \textbf{1.14} & \textbf{0.46} & \textbf{0.78} & \textbf{1.05} & \textbf{1.23} & \textbf{0.14} & \textbf{0.32} & \textbf{0.50} & \textbf{0.55} & \textbf{0.27} & \textbf{0.52} & \textbf{0.82} & \textbf{0.94}\\\hline
\end{tabular}
}
\label{tab:h36_short_ang}
\end{table}
\subsection{Datasets}
\noindent{{\bf Human3.6M}}~\cite{h36m_pami} is the most widely used benchmark dataset for motion prediction. It depicts seven actors performing 15 actions. Each human pose is represented as a 32-joint skeleton. We compute the 3D coordinates of the joints by applying forward kinematics on a standard skeleton as in~\cite{mao2019learning}. Following~\cite{LiZLL18,mao2019learning,Martinez_2017_CVPR}, we remove the global rotation, translation and constant angles or 3D coordinates of each human pose, and down-sample the motion sequences to 25 frames per second. As previous work~\cite{LiZLL18,mao2019learning,Martinez_2017_CVPR}, we test our method on subject 5 (S5). However, instead of testing on only 8 random sub-sequences per action, which was shown in~\cite{pavllo2019modeling} to lead to high variance, we report our results on 256 sub-sequences per action when using 3D coordinates. For fair comparison, we report our angular error on the same 8 sub-sequences used in~\cite{Tang_2018}. Nonetheless, we provide the angle-based results on 256 sub-sequences per action in the supplementary material.

\noindent{{\bf AMASS}}. The Archive of Motion Capture as Surface Shapes (AMASS) dataset~\cite{AMASS:2019} is a recently published human motion dataset, which unifies many mocap datasets, such as CMU, KIT and BMLrub, using a SMPL~\cite{SMPL:2015,MANO:SIGGRAPHASIA:2017} parameterization to obtain a human mesh. SMPL represents a human by a shape vector and joint rotation angles. The shape vector, which encompasses coefficients of different human shape bases, defines the human skeleton. We obtain human poses in 3D by applying forward kinematics to one human skeleton. In AMASS, a human pose is represented by 52 joints, including 22 body joints and 30 hand joints. Since we focus on predicting human body motion, we discard the hand joints and the 4 static joints, leading to an 18-joint human pose. As for H3.6M, we down-sample the frame-rate to 25Hz.
\begin{table}[ht]
\caption{Long-term prediction of joint angles on H3.6M.}
\centering
\resizebox{0.7\textwidth}{!}{%
\begin{tabular}{ccccc|cccc|cccc|cccc}
  & \multicolumn{4}{c}{Walking} & \multicolumn{4}{c}{Eating} & \multicolumn{4}{c}{Smoking} & \multicolumn{4}{c}{Discussion} \\
 milliseconds & 560 & 720 & 880 & 1000 & 560 & 720 & 880 & 1000 & 560 & 720 & 880 & 1000 & 560 & 720 & 880 & 1000\\\hline
convSeq2Seq~\cite{LiZLL18} & 0.87 & 0.96 & 0.97 & 1.00 & 0.86 & 0.90 & 1.12 & 1.24 & 0.98 & 1.11 & 1.42 & 1.67 & 1.42 & 1.76 & 1.90 & 2.03 \\ 
MHU~\cite{Tang_2018} & 1.44 & 1.46 & - & 1.44 & - & - & - & - & - & - & - & - & 1.37 & 1.66 & - & 1.88 \\
LTD-10-25~\cite{mao2019learning} & 0.65 & 0.69 & 0.69 & 0.67 & 0.76 & 0.82 & \textbf{1.00} & 1.12 & 0.87 & \textbf{0.99} & \textbf{1.33} & \textbf{1.57} & 1.33 & 1.53 & 1.62 & 1.70 \\
LTD-10-10~\cite{mao2019learning} & 0.69 & 0.77 & 0.76 & 0.77 & 0.76 & \textbf{0.81} & \textbf{1.00} & \textbf{1.10} & 0.88 & 1.01 & 1.36 & 1.58 & \textbf{1.27} & \textbf{1.51} & 1.66 & 1.75 \\\hline
Ours & \textbf{0.59} & \textbf{0.62} & \textbf{0.61} & \textbf{0.64} & \textbf{0.74} & \textbf{0.81} & 1.01 & \textbf{1.10} & \textbf{0.86} & 1.00 & 1.35 & 1.58 & 1.29 & \textbf{1.51} & \textbf{1.61} & \textbf{1.63} \\\hline
\end{tabular}
}
\resizebox{\textwidth}{!}{%
\begin{tabular}{ccccc|cccc|cccc|cccc|cccc|cccc}
  & \multicolumn{4}{c}{Directions} & \multicolumn{4}{c}{Greeting} & \multicolumn{4}{c}{Phoning} & \multicolumn{4}{c}{Posing}&\multicolumn{4}{c}{Purchases}&\multicolumn{4}{c}{Sitting} \\
  milliseconds& 560 & 720 & 880 & 1000 & 560 & 720 & 880 & 1000 & 560 & 720 & 880 & 1000 & 560 & 720 & 880 & 1000 & 560 & 720 & 880 & 1000 & 560 & 720 & 880 & 1000\\\hline
convSeq2Seq~\cite{LiZLL18} & 1.00 & 1.18 & 1.41 & 1.44 & 1.73 & 1.75 & 1.92 & 1.90 & 1.66 & 1.81 & 1.93 & 2.05 & 1.95 & 2.26 & 2.49 & 2.63 & 1.68 & 1.65 & 2.13 & 2.50 & 1.31 & 1.43 & 1.66 & 1.72\\
MHU~\cite{Tang_2018} & - & - & - & - & 1.75 & 1.74 & - & 1.87 & - & - & - & - & 1.82 & 2.17 & - & 2.51 & - & - & - & - & - & - & - & - \\
LTD-10-25~\cite{mao2019learning} & 0.84 & \textbf{1.02} & 1.23 & \textbf{1.26} & \textbf{1.43} & \textbf{1.44} & \textbf{1.59} & 1.59 & 1.45 & 1.57 & \textbf{1.66} & \textbf{1.65} & 1.62 & 1.94 & 2.22 & 2.42 & \textbf{1.42} & \textbf{1.48} & \textbf{1.93} & \textbf{2.21} & \textbf{1.08} & \textbf{1.20} & \textbf{1.39} & \textbf{1.45} \\
LTD-10-10~\cite{mao2019learning} & 0.90 & 1.07 & 1.32 & 1.35 & 1.47 & 1.47 & 1.63 & 1.59 & 1.49 & 1.64 & 1.75 & 1.74 & 1.61 & 2.02 & 2.35 & 2.55 & 1.47 & 1.57 & 1.99 & 2.27 & 1.12 & 1.25 & 1.46 & 1.52 \\\hline
Ours & \textbf{0.81} & \textbf{1.02} & \textbf{1.22} & 1.27 & 1.47 & 1.47 & 1.61 & \textbf{1.57} & \textbf{1.41} & \textbf{1.55} & 1.68 & 1.68 & \textbf{1.60} & \textbf{1.78} & \textbf{2.10} & \textbf{2.32} & 1.43 & 1.53 & 1.94 & 2.22 & 1.16 & 1.29 & 1.50 & 1.55 \\\hline
  & \multicolumn{4}{c}{Sitting Down} & \multicolumn{4}{c}{Taking Photo} & \multicolumn{4}{c}{Waiting} & \multicolumn{4}{c}{Walking Dog}&\multicolumn{4}{c}{Walking Together}&\multicolumn{4}{c}{Average} \\
  milliseconds & 560 & 720 & 880 & 1000 & 560 & 720 & 880 & 1000 & 560 & 720 & 880 & 1000 & 560 & 720 & 880 & 1000 & 560 & 720 & 880 & 1000 & 560 & 720 & 880 & 1000 \\\hline
convSeq2Seq~\cite{LiZLL18} & 1.45 & 1.70 & 1.85 & 1.98 & 1.09 & 1.18 & 1.27 & 1.32 & 1.68 & 2.02 & 2.33 & 2.45 & 1.73 & 1.85 & 1.99 & 2.04 & 0.82 & 0.89 & 0.95 & 1.29 & 1.35 & 1.50 & 1.69 & 1.82\\
MHU~\cite{Tang_2018} & - & - & - & - & 1.04 & 1.14 & - & 1.35 & - & - & - & - & 1.67 & 1.81 & - & 1.90 & - & - & - & - & 1.34 & 1.49 & 1.69 & 1.80 \\
LTD-10-25~\cite{mao2019learning} & 1.26 & 1.54 & 1.70 & 1.87 & 0.85 & 0.92 & 0.99 & 1.06 & 1.55 & \textbf{1.89} & \textbf{2.20} & \textbf{2.29} & 1.52 & \textbf{1.63} & 1.78 & 1.84 & 0.70 & 0.75 & 0.82 & \textbf{1.16} & 1.15 & 1.29 & 1.48 & 1.59 \\
LTD-10-10~\cite{mao2019learning} & \textbf{1.17} & \textbf{1.40} & \textbf{1.54} & \textbf{1.67} & \textbf{0.81} & \textbf{0.89} & \textbf{0.97} & \textbf{1.05} & 1.57 & 1.94 & 2.29 & 2.37 & 1.58 & 1.66 & 1.80 & 1.86 & 0.65 & 0.73 & 0.81 & \textbf{1.16} & 1.16 & 1.32 & 1.51 & 1.62 \\\hline
Ours & 1.18 & 1.42 & 1.55 & 1.70 & 0.82 & 0.91 & 1.00 & 1.08 & \textbf{1.54} & 1.90 & 2.22 & 2.30 & \textbf{1.57} & \textbf{1.63} & \textbf{1.76} & \textbf{1.82} & \textbf{0.63} & \textbf{0.68} & \textbf{0.79} & \textbf{1.16} & \textbf{1.14} & \textbf{1.28} & \textbf{1.46} & \textbf{1.57} \\\hline
\end{tabular}
}
\label{tab:h36_long_ang}
\end{table}
Since most sequences of the official testing split\footnote[1]{Described at https://github.com/nghorbani/amass} of AMASS consist of transition between two irrelevant actions, such as dancing to kicking, kicking to pushing, they are not suitable to evaluate our prediction algorithms, which assume that the history is relevant to forecast the future. Therefore, instead of using this official split, we treat BMLrub\footnote[2]{Available at https://amass.is.tue.mpg.de/dataset.} (522 min. video sequence), as our test set as each sequence consists of one actor performing one type of action. We then split the remaining parts of AMASS into training and validation data.

\noindent{{\bf 3DPW}}. The 3D Pose in the Wild dataset (3DPW)~\cite{vonMarcard2018} consists of challenging indoor and outdoor actions. We only evaluate our model trained on AMASS on the test set of 3DPW to show the generalization of our approach.

\begin{figure}[ht]
    \centering
    \includegraphics[width=\linewidth]{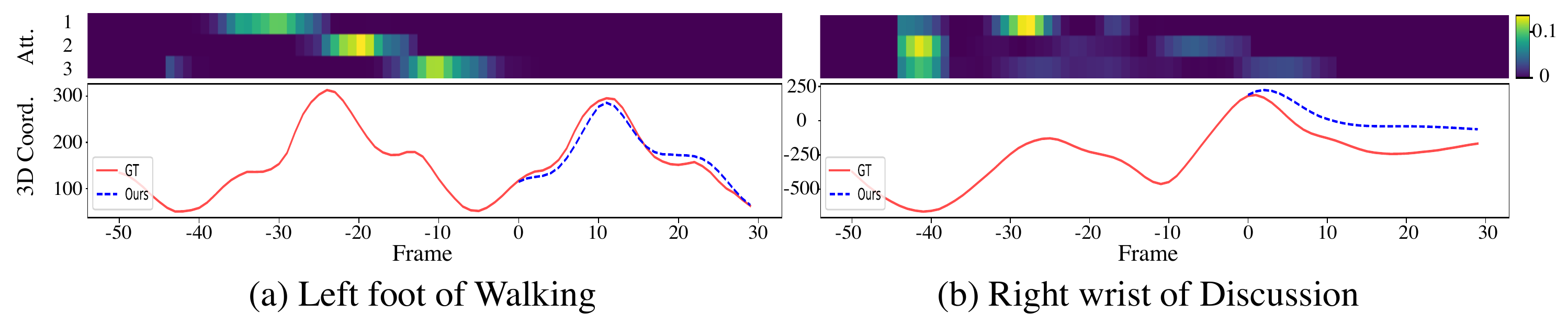}
    \caption{Visualization of attention maps and joint trajectories. The $x$-axis denotes the frame index, with prediction starting at frame 0. The $y$-axis of the attention map (top) is the prediction step. Specifically, since the model is trained to predict 10 future frames, we recursively perform prediction for 3 steps to generate 30 frames. Each row of an attention map is then the attention vector when predicting the corresponding 10 future frames. For illustration purpose, we show~\emph{per-frame} attention, which represents the attention for~\emph{its motion subsequence} consisting of M-1 frames forward and T frames afterwards. (a) Predicted attention map and trajectory of the left foot's $x$ coordinate for 'Walking', where the future motion closely resembles that between frames $-45$ and $-10$. Our model correctly attends to that very similar motion in the history. (b) Predicted attention map and trajectory of the right wrist's $x$ coordinate for 'Discussion'. In this case, the attention model searches for the most similar motion in the history. For example, in the $1^{st}$ prediction step, to predict frames $0$ to $10$ where a peak occurs, the model focuses on frames $-30$ to $-20$, where a similar peak pattern occurs.}
    \label{fig:att-h36m}
\end{figure}
\subsection{Evaluation Metrics and Baselines}
\noindent{{\bf Metrics}}. For the models that output 3D positions, we report the Mean Per Joint Position Error (MPJPE)~\cite{h36m_pami} in millimeter, which is commonly used in human pose estimation. For those that predict angles, we follow the standard evaluation protocol~\cite{Martinez_2017_CVPR,LiZLL18,mao2019learning} and report the Euclidean distance in Euler angle representation. 

\noindent{{\bf Baselines.}} We compare our approach with two RNN-based methods, Res. sup.~\cite{Martinez_2017_CVPR} and MHU~\cite{Tang_2018}, and two feed-forward models, convSeq2Seq~\cite{LiZLL18} and LTD~\cite{mao2019learning}, which constitutes the state of the art. The angular results of Res. sup.~\cite{Martinez_2017_CVPR}, convSeq2Seq~\cite{LiZLL18} and MHU on H3.6M are directly taken from the respective paper. For the other results of Res. sup.~\cite{Martinez_2017_CVPR} and convSeq2Seq~\cite{LiZLL18}, we adapt the code provided by the authors for H3.6M to 3D and AMASS. For LTD~\cite{mao2019learning}, we rely on the pre-trained models released by the authors for H3.6M, and train their model on AMASS using their official code. While Res. sup.~\cite{Martinez_2017_CVPR}, convSeq2Seq~\cite{LiZLL18} and MHU~\cite{Tang_2018} are all trained to generate 25 future frames, LTD~\cite{mao2019learning} has 3 different models, which we refer to as LTD-50-25~\cite{mao2019learning}, LTD-10-25~\cite{mao2019learning}, and LTD-10-10~\cite{mao2019learning}. The two numbers after the method name indicate the number of observed past frames and that of future frames to predict, respectively, during training. For example, LTD-10-25~\cite{mao2019learning} means that the model is trained to take the past 10 frames as input to predict the future 25 frames.

\subsection{Results}\label{sec:exp-res}
Following the setting of our baselines~\cite{Martinez_2017_CVPR,LiZLL18,Tang_2018,mao2019learning}, we report results for short-term ($< 500ms$) and long-term ($> 500ms$) prediction. On H3.6M, our model is trained using the past $50$ frames to predict the future $10$ frames, and we produce poses further in the future by recursively applying the predictions as input to the model. On AMASS, our model is trained using the past $50$ frames to predict the future $25$ frames.

\noindent{{\bf Human3.6M.}} In Tables~\ref{tab:h36_short_3d} and~\ref{tab:h36_long_3d}, we provide the H3.6M results for short-term and long-term prediction in 3D space, respectively. Note that we outperform all the baselines on average for both short-term and long-term prediction. In particular, our method yields larger improvements on activities with a clear repeated history, such as ``Walking" and ``Walking Together". Nevertheless, our approach remains competitive on the other actions. Note that we consistently outperform LTD-50-25, which is trained on the same number of past frames as our approach. This, we believe, evidences the benefits of exploiting attention on the motion history. 

\begin{table}[ht]
\caption{Short-term and long-term prediction of 3D joint positions on BMLrub (left) and 3DPW (right).}
\centering
\resizebox{0.9\textwidth}{!}{%
\begin{tabular}{ccccccccc||cccccccc}
& \multicolumn{8}{c}{AMASS-BMLrub} & \multicolumn{8}{c}{3DPW} \\
milliseconds   & 80   & 160  & 320  & 400  & 560 & 720 & 880 & 1000 & 80   & 160  & 320  & 400  & 560 & 720 & 880 & 1000 \\\hline
convSeq2Seq~\cite{LiZLL18} & 20.6 & 36.9 & 59.7 & 67.6 & 79.0 & 87.0 & 91.5 & 93.5 & 18.8 & 32.9 & 52.0 & 58.8 & 69.4 & 77.0 & 83.6 & 87.8 \\
LTD-10-10~\cite{mao2019learning} & \textbf{10.3} & \textbf{19.3} & 36.6 & 44.6 & 61.5 & 75.9 & 86.2 & 91.2 & \textbf{12.0} & \textbf{22.0} & \textbf{38.9} & 46.2 & 59.1 & 69.1 & 76.5 & 81.1\\
LTD-10-25~\cite{mao2019learning} & 11.0 & 20.7 & 37.8 & 45.3 & 57.2 & 65.7 & 71.3 & 75.2 & 12.6 & 23.2 & 39.7 & 46.6 & 57.9 & 65.8 & 71.5 & 75.5 \\\hline
Ours & 11.3 & 20.7 & \textbf{35.7} & \textbf{42.0} & \textbf{51.7} & \textbf{58.6} & \textbf{63.4} & \textbf{67.2} & 12.6 & 23.1 & 39.0 & \textbf{45.4} & \textbf{56.0} & \textbf{63.6} & \textbf{69.7} & \textbf{73.7} \\\hline
\end{tabular}
}
\label{tab:amass_3d}
\end{table}
Let us now focus on the LTD~\cite{mao2019learning} baseline, which constitutes the state of the art. Although LTD-10-10 is very competitive for short-term prediction, when it comes to generate poses in the further future, it yields higher average error, i.e., $114.0mm$ at $1000ms$. By contrast, LTD-10-25 and LTD-50-25 achieve good performance at $880ms$ and above, but perform worse than LTD-10-10 at other time horizons. Our approach, however, yields state-of-the-art performance for both short-term and long-term predictions. To summarize, our motion attention model improves the performance of the predictor for short-term prediction and further enables it to generate better long-term predictions. This is further evidenced by Tables~\ref{tab:h36_short_ang} and~\ref{tab:h36_long_ang}, where we report the short-term and long-term prediction results in angle space on H3.6M, and by the qualitative comparison in Fig.~\ref{fig:short-pred-quali-h36m}. More qualitative results are provided in the supplementary material.

\noindent{{\bf AMASS \& 3DPW.}} The results of short-term and long-term prediction in 3D on AMASS and 3DPW are shown in Table~\ref{tab:amass_3d}. Our method consistently outperforms baseline approaches, which further evidences the benefits of our motion attention model. Since none of the methods were trained on 3DPW, these results further demonstrate that our approach generalizes better to new datasets than the baselines.

\noindent\textbf{Visualisation of attention.} In Fig.~\ref{fig:att-h36m}, we visualize the attention maps computed by our motion attention model on a few sampled joints for their corresponding coordinate trajectories. In particular, we show attention maps for joints in a periodical motion (``Walking") and a non-periodical one (``Discussion"). In both cases, the attention model can find the most relevant sub-sequences in the history, which encode either a nearly identical motion (periodical action), or a similar pattern (non-periodical action).

\noindent{{\bf Motion repeats itself in longer-term history.}} Our model, which is trained with fixed-length observations, can nonetheless exploit longer history at test time if it is available. To evaluate this and our model's ability to capture long-range motion dependencies, we manually sampled $100$ sequences from the test set of H3.6M, in which similar motion occurs in the further past than that used to train our model.

\begin{figure}[ht]
    \centering
      \includegraphics[width=\textwidth]{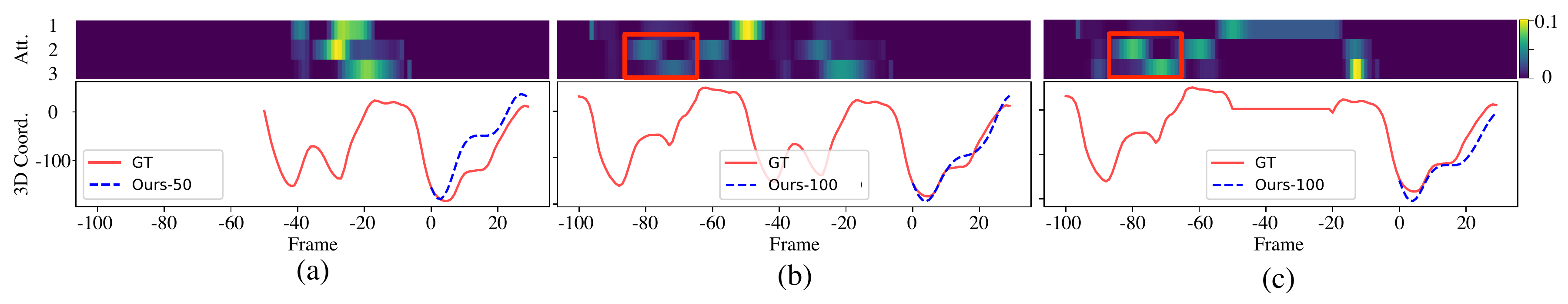}
      \caption{Visualization of attention maps and joint coordinate trajectories for ``Smoking" on H3.6M. (a) Results of our model observing 50 past frames. (b) Results of our model observing 100 frames. (c) Results obtained when replacing the motion of the past 40 frames with a constant pose.}
      \label{fig:his_rep_att}
\end{figure}

In Table~\ref{tab:his_rep}, we compare the results of a model trained with 50 past frames and using either $50$ frames (Ours-50) or 100 frames (Ours-100) at test time. Although the performance is close in the very short term ($<160ms$), the benefits of our model using longer history become obvious when it comes to further future, leading to a performance boost of $4.2mm$ at $1s$. In Fig.~\ref{fig:his_rep_att}, we compare the attention maps and predicted joint trajectories of Ours-50 (a) and Ours-100 (b). The highlighted regions (in red box) in the attention map demonstrate that our model can capture the repeated motions in the further history if it is available during test and improve the motion prediction results.

\begin{table}[ht]
\caption{Short-term and long-term prediction of 3D positions on selected sequences where similar patterns occur in the longer history. The number after ``Ours" indicates the observed frames during testing. Both methods observed 50 frames during training.}
\centering
\resizebox{0.5\textwidth}{!}{%
\begin{tabular}{ccccccccc}
milliseconds   & 80   & 160  & 320  & 400  & 560 & 720 & 880 & 1000 \\\hline
Ours-50 & \textbf{10.7} & \textbf{22.4} & 46.9 & 58.3 & 79.0 & 97.1 & 111.0 & 121.1\\
Ours-100 & \textbf{10.7} & 22.5 & \textbf{46.4} & \textbf{57.5} & \textbf{77.8} & \textbf{95.1} & \textbf{107.6} & \textbf{116.9}\\\hline
\end{tabular}
}
\label{tab:his_rep}
\end{table}
To show the influence of further historical frames, we replace the past $40$ frames with a static pose, thus removing the motion in that period, and then perform prediction with this sequence. As shown in~Fig.~\ref{fig:his_rep_att} (c), attending to the similar motion between frames $-80$ and $-60$, yields a trajectory much closer to the ground truth than only attending to the past $50$ frames.

\section{Conclusion}
In this paper, we have introduced an attention-based motion prediction approach that selectively exploits historical information according to the similarity between the current motion context and the sub-sequences in the past. This has led to a predictor equipped with a~\emph{motion attention} model that can effectively make use of historical motions, even when they are far in the past. Our approach achieves state-of-the-art performance on the commonly-used motion prediction benchmarks and on recently-published datasets. Furthermore, our experiments have demonstrated that our network generalizes to previously-unseen datasets without re-training or fine-tuning, and can handle longer history than that it was trained with to further boost performance on non-periodical motions with repeated history. In the future, we will further investigate the use of our motion attention mechanisms to discover human motion patterns in body parts level such as legs and arms to get more flexible attentions and explore new prediction frame works.
\section*{Acknowledgements}
This research was supported in part by the Australia Research Council DECRA Fellowship (DE180100628) and ARC Discovery Grant (DP200102274). The authors would like to thank NVIDIA for the donated GPU (Titan V).

\newpage
\setcounter{section}{0}
\setcounter{figure}{0}
\setcounter{table}{0}
\pagestyle{headings}
\mainmatter
\def\ECCVSubNumber{2186}  

\title{History Repeats Itself: Human Motion Prediction via Motion Attention \\-----Supplementary Material-----} 

\titlerunning{History Repeats Itself: Human Motion Prediction via Motion Attention}
%
\author{Wei Mao\inst{1} \and
	Miaomiao Liu\inst{1} \and
	Mathieu Salzmann\inst{2}}
\authorrunning{W. Mao, M. Liu, M. Salzmann}
%
\institute{Australian National University, Canberra, Australia
	\and
	EPFL--CVLab \& ClearSpace, Switzerland\\
	\email{\{wei.mao,miaomiao.liu\}@anu.edu.au, mathieu.salzmann@epfl.ch}
}
\maketitle

\section{Datasets}
Below we provide more details about the datasets used in our experiments.

\noindent{\textbf{Human3.6M.}} As in~\cite{mao2019learning}, we use the skeleton of the subject 1 (S1) of Human3.6M as standard skeleton to compute the 3D joint coordinates from the joint angle representation. After removing the global rotation, translation and constant angles or 3D
coordinates of each human pose, this leaves us with a 48 dimensional vector and a 66 dimensional vector for human pose in angle representation and 3D position, respectively. As in~\cite{mao2019learning,LiZLL18,Martinez_2017_CVPR}, the rotation angles are represented as exponential maps. During training, we set aside subject 11 (S11) as our validation set to choose the model that achieves the best performance across all future frames, and the remaining 5 subjects (S1,S6,S7,S8,S9) are used as training set.

\noindent{\textbf{AMASS \& 3DPW.}} The human skeleton in AMASS and 3DPW is defined by a shape vector. In our experiment, we obtain the 3D joint positions by applying forward kinematic on the skeleton derived from the shape vector of the CMU dataset.
\begin{table}[t]
\caption{Short-term prediction of joint angles on H3.6M. We report the results on 256 sub-sequences per action.}
\centering
\resizebox{0.9\textwidth}{!}{%
\begin{tabular}{ccccc|cccc|cccc|cccc}
  & \multicolumn{4}{c}{Walking} & \multicolumn{4}{c}{Eating} & \multicolumn{4}{c}{Smoking} & \multicolumn{4}{c}{Discussion} \\
     milliseconds       & 80   & 160  & 320  & 400  & 80   & 160  & 320  & 400  & 80   & 160  & 320  & 400  & 80   & 160  & 320  & 400   \\\hline
LTD-10-25~\cite{mao2019learning}&0.26 & 0.47 & 0.73 & 0.80 & 0.21 & 0.45 & 0.71 & 0.82 & 0.26 & 0.43 & 0.74 & 0.86 & 0.48 & \textbf{0.67} & 1.10 & 1.28 \\
LTD-10-10~\cite{mao2019learning}&0.25 & 0.45 & 0.72 & 0.78 & \textbf{0.20} & \textbf{0.41} & 0.70 & 0.82 & 0.25 & 0.41 & \textbf{0.71} & \textbf{0.83} & 0.47 & 0.68 & \textbf{1.09} & \textbf{1.25} \\\hline
Ours&\textbf{0.24} & \textbf{0.43} & \textbf{0.66} & \textbf{0.71} & \textbf{0.20} & \textbf{0.41} & \textbf{0.68} & \textbf{0.80} & \textbf{0.25} & \textbf{0.41} & \textbf{0.71} & \textbf{0.83} & \textbf{0.44} & 0.68 & \textbf{1.09} & \textbf{1.25}\\\hline
\end{tabular}
}
\resizebox{\textwidth}{!}{%
\begin{tabular}{ccccc|cccc|cccc|cccc|cccc|cccc}
  & \multicolumn{4}{c}{Directions} & \multicolumn{4}{c}{Greeting} & \multicolumn{4}{c}{Phoning} & \multicolumn{4}{c}{Posing}&\multicolumn{4}{c}{Purchases}&\multicolumn{4}{c}{Sitting} \\
  milliseconds& 80 & 160 & 320 & 400 & 80 & 160 & 320 & 400 & 80 & 160 & 320 & 400 & 80 & 160 & 320 & 400& 80 & 160 & 320 & 400& 80 & 160 & 320 & 400 \\ \hline
LTD-10-25~\cite{mao2019learning}&0.20 & 0.41 & 0.76 & 0.92 & 0.52 & 0.84 & 1.24 & 1.41 & 0.34 & 0.57 & 0.96 & 1.09 & 0.31 & 0.60 & 1.06 & 1.24 & 0.47 & 0.84 & 1.24 & 1.33 & 0.33 & 0.52 & 0.92 & 1.06 \\
LTD-10-10~\cite{mao2019learning}&\textbf{0.19} & 0.39 & 0.75 & 0.91 & 0.53 & 0.82 & 1.22 & 1.39 & 0 .33 & \textbf{0.54} & \textbf{0.94} & \textbf{1.07} & 0.30 & 0.61 & 1.02 & \textbf{1.20} & 0.45 & 0.80 & 1.22 & \textbf{1.32} & 0.28 & \textbf{0.56} & \textbf{0.94} & 1.08 \\\hline
Ours&\textbf{0.19} & \textbf{0.38} & \textbf{0.74} & \textbf{0.90} & \textbf{0.50} & \textbf{0.79} & \textbf{1.21} & \textbf{1.38} & \textbf{0.32} & \textbf{0.54} & \textbf{0.94} & \textbf{1.07} & \textbf{0.27} & \textbf{0.57} & \textbf{1.00} & 1.22 & \textbf{0.43} & \textbf{0.79} & \textbf{1.21} & \textbf{1.32} & \textbf{0.27} & \textbf{0.56} & \textbf{0.94} & \textbf{1.06}\\\hline
  & \multicolumn{4}{c}{Sitting Down} & \multicolumn{4}{c}{Taking Photo} & \multicolumn{4}{c}{Waiting} & \multicolumn{4}{c}{Walking Dog}&\multicolumn{4}{c}{Walking Together}&\multicolumn{4}{c}{Average} \\
  milliseconds& 80 & 160 & 320 & 400 & 80 & 160 & 320 & 400 & 80 & 160 & 320 & 400 & 80 & 160 & 320 & 400& 80 & 160 & 320 & 400& 80 & 160 & 320 & 400 \\ \hline
LTD-10-25~\cite{mao2019learning}&0.44 & 0.75 & 1.21 & 1.40 & 0.21 & 0.35 & 0.62 & 0.74 & 0.29 & 0.49 & 0.92 & 1.07 & 0.44 & 0.71 & 1.04 & 1.14 & 0.26 & 0.43 & 0.67 & 0.77 & 0.34 & 0.57 & 0.93 & 1.06 \\
LTD-10-10~\cite{mao2019learning}&\textbf{0.43} & \textbf{0.74} & \textbf{1.20} & \textbf{1.38} & 0.20 & \textbf{0.34} & 0.61 & \textbf{0.72} & 0.28 & \textbf{0.47} & \textbf{0.90} & \textbf{1.05} & 0.43 & 0.69 & 1.02 & 1.13 & \textbf{0.24} & 0.40 & 0.63 & 0.73 & 0.32 & \textbf{0.55} & 0.91 & \textbf{1.04} \\\hline
Ours&\textbf{0.43} & \textbf{0.74} & \textbf{1.20} & 1.39 & \textbf{0.19} & \textbf{0.34} & \textbf{0.60} & \textbf{0.72} & \textbf{0.27} & \textbf{0.47} & 0.91 & 1.07 & \textbf{0.42} & \textbf{0.68} & \textbf{1.01} & \textbf{1.12} & \textbf{0.24} & \textbf{0.39} & \textbf{0.62} & \textbf{0.71} & \textbf{0.31} & \textbf{0.55} & \textbf{0.90} & \textbf{1.04}\\\hline
\end{tabular}
}
\label{tab:supp_h36_short_ang}
\end{table}
\begin{table}[t]
\caption{Long-term prediction of joint angles on H3.6M.}
\centering
\resizebox{0.9\textwidth}{!}{%
\begin{tabular}{ccccc|cccc|cccc|cccc}
  & \multicolumn{4}{c}{Walking} & \multicolumn{4}{c}{Eating} & \multicolumn{4}{c}{Smoking} & \multicolumn{4}{c}{Discussion} \\
 milliseconds & 560 & 720 & 880 & 1000 & 560 & 720 & 880 & 1000 & 560 & 720 & 880 & 1000 & 560 & 720 & 880 & 1000\\\hline
LTD-10-25~\cite{mao2019learning}&0.92 & 0.97 & 1.03 & 1.05 & 0.99 & 1.16 & 1.26 & 1.33 & 1.07 & 1.26 & 1.41 & 1.55 & 1.48 & \textbf{1.59} & \textbf{1.68} & \textbf{1.76} \\
LTD-10-10~\cite{mao2019learning}&0.95 & 1.03 & 1.09 & 1.12 & \textbf{0.98} & 1.15 & 1.28 & 1.36 & 1.04 & 1.21 & \textbf{1.36} & 1.51 & \textbf{1.47} & \textbf{1.59} & 1.71 & 1.79 \\\hline
Ours & \textbf{0.84} & \textbf{0.91} & \textbf{0.99} & \textbf{1.03} & \textbf{0.98} & \textbf{1.14} & \textbf{1.24} & \textbf{1.31} & \textbf{1.04} & \textbf{1.20} & 1.38 & \textbf{1.50} & 1.49 & 1.62 & 1.72 & 1.82
 \\\hline
\end{tabular}
}

\resizebox{\textwidth}{!}{%
\begin{tabular}{ccccc|cccc|cccc|cccc|cccc|cccc}
  & \multicolumn{4}{c}{Directions} & \multicolumn{4}{c}{Greeting} & \multicolumn{4}{c}{Phoning} & \multicolumn{4}{c}{Posing}&\multicolumn{4}{c}{Purchases}&\multicolumn{4}{c}{Sitting} \\
  milliseconds& 560 & 720 & 880 & 1000 & 560 & 720 & 880 & 1000 & 560 & 720 & 880 & 1000 & 560 & 720 & 880 & 1000 & 560 & 720 & 880 & 1000 & 560 & 720 & 880 & 1000\\\hline
LTD-10-25~\cite{mao2019learning}&1.10 & 1.23 & 1.35 & \textbf{1.41} & 1.63 & 1.81 & 1.95 & 2.01 & 1.29 & \textbf{1.48} & \textbf{1.63} & \textbf{1.74} & \textbf{1.54} & 1.81 & \textbf{2.10} & \textbf{2.23} & 1.51 & 1.66 & 1.80 & 1.87 & 1.34 & 1.60 & \textbf{1.79} & \textbf{1.87} \\
LTD-10-10~\cite{mao2019learning}&1.09 & \textbf{1.21} & \textbf{1.34} & \textbf{1.41} & 1.63 & 1.82 & 1.99 & 2.06 & 1.29 & 1.50 & 1.67 & 1.78 & 1.53 & 1.81 & 2.12 & 2.25 & 1.52 & 1.68 & 1.83 & 1.91 & 1.34 & 1.60 & \textbf{1.79} & 1.89 \\\hline
Ours& \textbf{1.08} & 1.22 & 1.35 & 1.42 & \textbf{1.62} & \textbf{1.79} & \textbf{1.93} & \textbf{1.99} & \textbf{1.28} & 1.49 & 1.65 & 1.76 & 1.55 & \textbf{1.80} & \textbf{2.10} & 2.24 & \textbf{1.47} & \textbf{1.62} & \textbf{1.75} & \textbf{1.82} & \textbf{1.33} & \textbf{1.59} & \textbf{1.79} & 1.88
 \\\hline
  & \multicolumn{4}{c}{Sitting Down} & \multicolumn{4}{c}{Taking Photo} & \multicolumn{4}{c}{Waiting} & \multicolumn{4}{c}{Walking Dog}&\multicolumn{4}{c}{Walking Together}&\multicolumn{4}{c}{Average} \\
  milliseconds & 560 & 720 & 880 & 1000 & 560 & 720 & 880 & 1000 & 560 & 720 & 880 & 1000 & 560 & 720 & 880 & 1000 & 560 & 720 & 880 & 1000 & 560 & 720 & 880 & 1000 \\\hline
LTD-10-25~\cite{mao2019learning}&1.71 & 1.95 & 2.17 & 2.26 & 0.94 & 1.10 & 1.23 & 1.34 & \textbf{1.30} & 1.48 & \textbf{1.63} & \textbf{1.74} & \textbf{1.30} & \textbf{1.45} & \textbf{1.55} & 1.64 & 0.91 & 0.98 & 1.02 & 1.06 & 1.27 & 1.44 & 1.57 & 1.66 \\
LTD-10-10~\cite{mao2019learning}&\textbf{1.68} & 1.91 & \textbf{2.13} & \textbf{2.22} & 0.93 & 1.08 & 1.22 & 1.34 & \textbf{1.30} & \textbf{1.47} & \textbf{1.63} & 1.75 & 1.31 & 1.48 & 1.59 & 1.68 & 0.89 & 0.98 & 1.03 & 1.08 & 1.26 & 1.44 & 1.59 & 1.68 \\\hline
Ours & \textbf{1.68} & \textbf{1.90} & \textbf{2.12} & \textbf{2.22} & \textbf{0.92} & \textbf{1.07} & \textbf{1.21} & \textbf{1.33} & 1.31 & 1.49 & 1.64 & 1.77 & \textbf{1.30} & \textbf{1.45} & \textbf{1.55} & \textbf{1.63} & \textbf{0.86} & \textbf{0.94} & \textbf{1.00} & \textbf{1.04} & \textbf{1.25} & \textbf{1.42} & \textbf{1.56} & \textbf{1.65}
 \\\hline
\end{tabular}
}
\label{tab:supp_h36_long_ang}
\end{table}
As specified in the main paper, we evaluate the model on BMLrub and 3DPW. Each video sequence is first downsampled to 25 frames per second, and evaluate on sub-sequences of length $M+T$ that start from every $5^{th}$ frame of each video sequence.
\section{Implementation Details}
 We implemented our network in Pytorch~\cite{paszke2017automatic} and trained it using the ADAM optimizer~\cite{kingma2014adam}. We use a learning rate of $0.0005$ with a decay at every epoch so as to make the learning rate be $0.00005$ at the $50^{th}$ epoch. We train our model for $50$ epochs with a batch size of 32 for H3.6M and 128 for AMASS. One forward and backward pass takes 32ms for H3.6M and 45ms for AMASS on an NVIDIA Titan V GPU.
\section{Additional Results on H3.6M}
\subsection{Results on 256 Random Sub-sequences}
In Table~\ref{tab:supp_h36_short_ang} and \ref{tab:supp_h36_long_ang}, we report the Human3.6M results in angle representation for short-term and long-term prediction, respectively. Here, we average the error over 256 random sub-sequences per action, which was proven in~\cite{pavllo2019modeling} to be more stable than averaging over 8 random sub-sequences per action as is commonly done. Our conclusions remain unchanged: our approach achieves the state-of-the-art performance for both short-term and long-term prediction on average.

\subsection{Generating Long Future for Periodical Motions}
For periodical motions, such as ``Walking", our approach can generate very long futures (up to 16 seconds). As shown in the supplementary video, such future predictions are hard to distinguish from the ground truth even for humans.

\section{Additional Results on AMASS}
\begin{figure}[ht]
    \centering
    \begin{tabular}{c}
      \includegraphics[width=\linewidth]{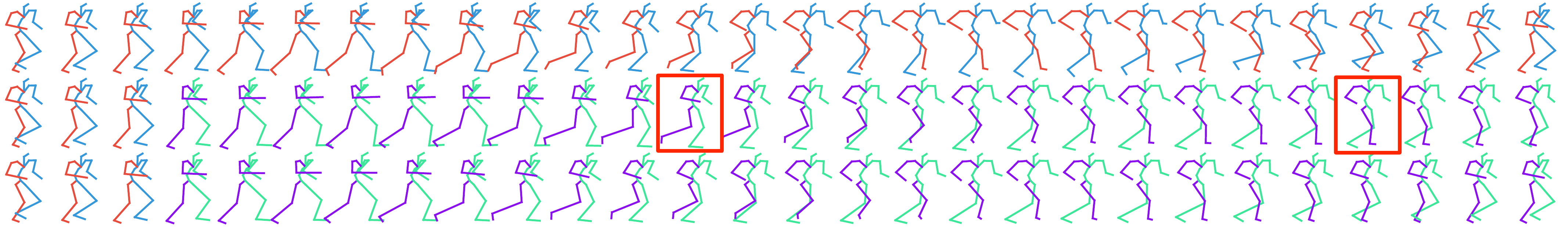} \\
      (a) Jogging\\
      \includegraphics[width=\linewidth]{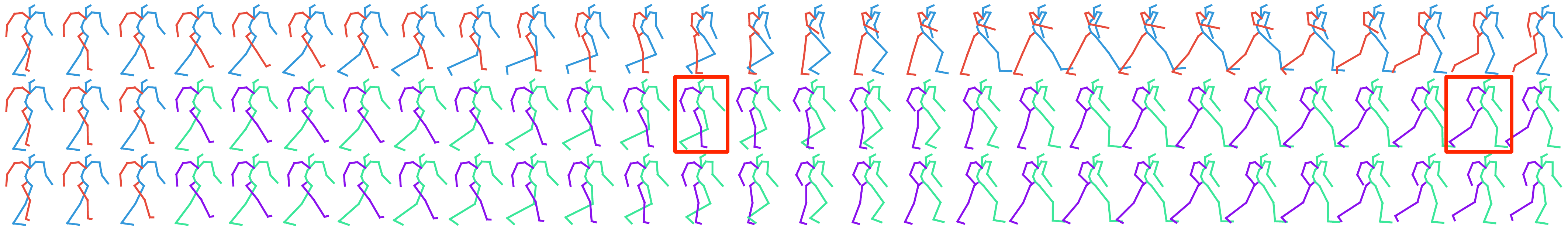} \\
      (b) Walking\\
      \includegraphics[width=\linewidth]{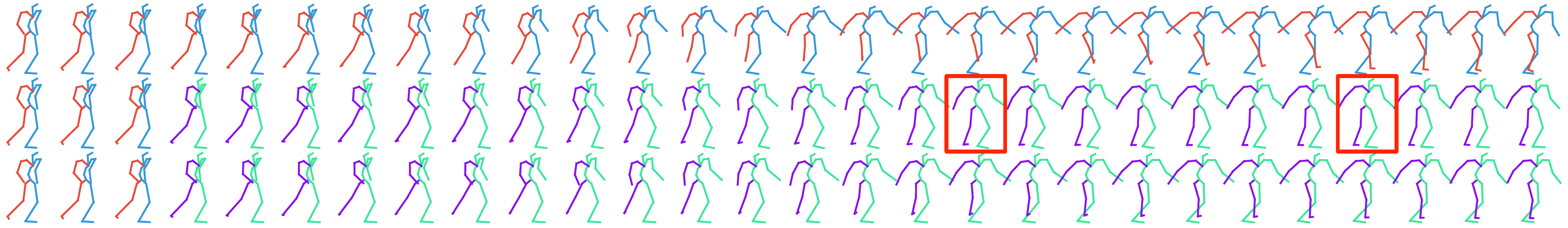} \\
      (c) Stretching
    \end{tabular}
    \caption{Qualitative comparison on the BMLrub dataset. From top to bottom, we show the ground-truth motion, the prediction results of LTD~\cite{mao2019learning} and of our approach on 3D position. The observed poses are shown as blue and red skeletons and the predictions in green and purple. As highlighted by the red boxes, our predictions better match the ground truth, in particular for the legs.}
    \label{fig:vis_amass}
\end{figure}

In Fig.~\ref{fig:vis_amass}, we compare the results of LTD~\cite{mao2019learning} and of our approach on the BMLrub dataset. Our results better match the ground truth.

\section{Motion Attention vs. Frame-wise Attention}
To further investigate the influence of \emph{motion} attention, where the attention on the history sub-sequences $\{\textbf{X}_{i:i+M+T-1}\}_{i=1}^{N-M-T+1}$ is a function of the first $M$ poses of every sub-sequence $\{\textbf{X}_{i:i+M-1}\}_{i=1}^{N-M-T+1}$ (keys) and the last observed $M$ poses $\textbf{X}_{N-M+1:N}$ (query), we replace the keys and query with the last frame of each sub-sequence. That is, we use $\{\textbf{X}_{i+M-1}\}_{i=1}^{N-M-T+1}$ as keys and $\textbf{X}_{N}$ as query. We refer to the resulting method as \emph{Frame-wise Attention}. As shown in Table~\ref{tab:ablation}, motion attention outperforms frame-wise attention by a large margin. As discussed in the main paper, this is due to frame-wise attention not considering the direction of the motion, leading to ambiguities.
\begin{table}[ht]
\caption{Comparison of frame-wise attention and with our motion attention.}
\centering
\resizebox{0.8\textwidth}{!}{%
\begin{tabular}{ccccccccc}
milliseconds   & 80   & 160  & 320  & 400  & 560 & 720 & 880 & 1000 \\\hline
Frame-wise Attention & 24.0 & 44.5 & 76.1 & 88.3 & 107.5 & 121.7 & 131.7 & 136.7\\
Motion Attention & \textbf{10.8} & \textbf{23.9} & \textbf{49.4} & \textbf{60.7} & \textbf{77.3} & \textbf{92.0} & \textbf{104.4} & \textbf{112.4}\\\hline
\end{tabular}
}
\label{tab:ablation}
\end{table}

\clearpage
%
%
\bibliographystyle{splncs04}
\bibliography{egbib}

\begin{thebibliography}{10}
\providecommand{\url}[1]{\texttt{#1}}
\providecommand{\urlprefix}{URL }
\providecommand{\doi}[1]{https://doi.org/#1}

\bibitem{ArjovskyB17}
Arjovsky, M., Bottou, L.: Towards principled methods for training generative
  adversarial networks. In: ICLR (2017)

\bibitem{bahdanau2014neural}
Bahdanau, D., Cho, K., Bengio, Y.: Neural machine translation by jointly
  learning to align and translate (2015)

\bibitem{brand2000style}
Brand, M., Hertzmann, A.: Style machines. In: Proceedings of the 27th annual
  conference on Computer graphics and interactive techniques. pp. 183--192. ACM
  Press/Addison-Wesley Publishing Co. (2000)

\bibitem{Butepage_2017_CVPR}
Butepage, J., Black, M.J., Kragic, D., Kjellstrom, H.: Deep representation
  learning for human motion prediction and classification. In: CVPR (July 2017)

\bibitem{fragkiadaki2015recurrent}
Fragkiadaki, K., Levine, S., Felsen, P., Malik, J.: Recurrent network models
  for human dynamics. In: ICCV. pp. 4346--4354 (2015)

\bibitem{gong2011multi}
Gong, H., Sim, J., Likhachev, M., Shi, J.: Multi-hypothesis motion planning for
  visual object tracking. In: ICCV. pp. 619--626. IEEE (2011)

\bibitem{gopalakrishnan2019neural}
Gopalakrishnan, A., Mali, A., Kifer, D., Giles, L., Ororbia, A.G.: A neural
  temporal model for human motion prediction. In: CVPR. pp. 12116--12125 (2019)

\bibitem{gui2018adversarial}
Gui, L.Y., Wang, Y.X., Liang, X., Moura, J.M.: Adversarial geometry-aware human
  motion prediction. In: ECCV. pp. 786--803 (2018)

\bibitem{hernandez2019human}
Hernandez, A., Gall, J., Moreno-Noguer, F.: Human motion prediction via
  spatio-temporal inpainting. In: ICCV. pp. 7134--7143 (2019)

\bibitem{h36m_pami}
Ionescu, C., Papava, D., Olaru, V., Sminchisescu, C.: Human3.6m: Large scale
  datasets and predictive methods for 3d human sensing in natural environments.
  TPAMI  \textbf{36}(7),  1325--1339 (jul 2014)

\bibitem{JainZSS16}
Jain, A., Zamir, A.R., Savarese, S., Saxena, A.: Structural-rnn: Deep learning
  on spatio-temporal graphs. In: CVPR. pp. 5308--5317 (2016)

\bibitem{kingma2014adam}
Kingma, D.P., Ba, J.: Adam: A method for stochastic optimization. ICLR  (2015)

\bibitem{kiros2015skip}
Kiros, R., Zhu, Y., Salakhutdinov, R.R., Zemel, R., Urtasun, R., Torralba, A.,
  Fidler, S.: Skip-thought vectors. In: NIPS. pp. 3294--3302 (2015)

\bibitem{koppula2013anticipating}
Koppula, H.S., Saxena, A.: Anticipating human activities for reactive robotic
  response. In: IROS. p.~2071. Tokyo (2013)

\bibitem{kovar2008motion}
Kovar, L., Gleicher, M., Pighin, F.: Motion graphs. In: ACM SIGGRAPH 2008
  classes, pp. 1--10 (2008)

\bibitem{2012-ccclde}
Levine, S., Wang, J.M., Haraux, A., Popovi\'{c}, Z., Koltun, V.: Continuous
  character control with low-dimensional embeddings. ACM Transactions on
  Graphics  \textbf{31}(4), ~28 (2012)

\bibitem{LiZLL18}
Li, C., Zhang, Z., Lee, W.S., Lee, G.H.: Convolutional sequence to sequence
  model for human dynamics. In: CVPR. pp. 5226--5234 (2018)

\bibitem{SMPL:2015}
Loper, M., Mahmood, N., Romero, J., Pons-Moll, G., Black, M.J.: {SMPL}: A
  skinned multi-person linear model. ACM Trans. Graphics (Proc. SIGGRAPH Asia)
  \textbf{34}(6),  248:1--248:16 (Oct 2015)

\bibitem{AMASS:2019}
Mahmood, N., Ghorbani, N., Troje, N.F., Pons-Moll, G., Black, M.J.: Amass:
  Archive of motion capture as surface shapes. In: ICCV (Oct 2019),
  \url{https://amass.is.tue.mpg.de}

\bibitem{mao2019learning}
Mao, W., Liu, M., Salzmann, M., Li, H.: Learning trajectory dependencies for
  human motion prediction. In: ICCV. pp. 9489--9497 (2019)

\bibitem{vonMarcard2018}
von Marcard, T., Henschel, R., Black, M., Rosenhahn, B., Pons-Moll, G.:
  Recovering accurate 3d human pose in the wild using imus and a moving camera.
  In: ECCV (sep 2018)

\bibitem{Martinez_2017_CVPR}
Martinez, J., Black, M.J., Romero, J.: On human motion prediction using
  recurrent neural networks. In: CVPR (July 2017)

\bibitem{nair2010rectified}
Nair, V., Hinton, G.E.: Rectified linear units improve restricted boltzmann
  machines. In: ICML. pp. 807--814 (2010)

\bibitem{paszke2017automatic}
Paszke, A., Gross, S., Chintala, S., Chanan, G., Yang, E., DeVito, Z., Lin, Z.,
  Desmaison, A., Antiga, L., Lerer, A.: Automatic differentiation in pytorch.
  In: NIPS-W (2017)

\bibitem{pavllo2019modeling}
Pavllo, D., Feichtenhofer, C., Auli, M., Grangier, D.: Modeling human motion
  with quaternion-based neural networks. IJCV pp. 1--18 (2019)

\bibitem{MANO:SIGGRAPHASIA:2017}
Romero, J., Tzionas, D., Black, M.J.: Embodied hands: Modeling and capturing
  hands and bodies together. ACM Transactions on Graphics, (Proc. SIGGRAPH
  Asia)  \textbf{36}(6) (Nov 2017)

\bibitem{runia2018real}
Runia, T.F., Snoek, C.G., Smeulders, A.W.: Real-world repetition estimation by
  div, grad and curl. In: CVPR. pp. 9009--9017 (2018)

\bibitem{sidenbladh2002implicit}
Sidenbladh, H., Black, M.J., Sigal, L.: Implicit probabilistic models of human
  motion for synthesis and tracking. In: ECCV. pp. 784--800. Springer (2002)

\bibitem{sutskever2011generating}
Sutskever, I., Martens, J., Hinton, G.E.: Generating text with recurrent neural
  networks. In: ICML. pp. 1017--1024 (2011)

\bibitem{Tang_2018}
Tang, Y., Ma, L., Liu, W., Zheng, W.S.: Long-term human motion prediction by
  modeling motion context and enhancing motion dynamics. IJCAI  (Jul 2018).
  \doi{10.24963/ijcai.2018/130},
  \url{http://dx.doi.org/10.24963/ijcai.2018/130}

\bibitem{vaswani2017attention}
Vaswani, A., Shazeer, N., Parmar, N., Uszkoreit, J., Jones, L., Gomez, A.N.,
  Kaiser, {\L}., Polosukhin, I.: Attention is all you need. In: NIPS. pp.
  5998--6008 (2017)

\bibitem{wang2008gaussian}
Wang, J.M., Fleet, D.J., Hertzmann, A.: Gaussian process dynamical models for
  human motion. TPAMI  \textbf{30}(2),  283--298 (2008)

\end{thebibliography}
\end{document}